\documentclass[10pt,twocolumn,letterpaper]{article}
\usepackage[accsupp]{axessibility}
\usepackage[pagenumbers]{cvpr} %

\definecolor{cvprblue}{rgb}{0.21,0.49,0.74}
\usepackage[pagebackref,breaklinks,colorlinks,allcolors=cvprblue]{hyperref}
\usepackage{booktabs}
\usepackage{multirow}
\usepackage{graphicx}
\usepackage[dvipsnames]{xcolor}

\def\paperID{*****}
\def\confName{CVPR}
\def\confYear{2026}

\title{Towards Robust Sequential Decomposition for Complex Image Editing}

\author{
Zilai Zeng\textsuperscript{1,2,}\thanks{Work done when ZZ was an intern at ByteDance} \quad Mingdeng Cao\textsuperscript{3} \quad Zijie Li\textsuperscript{2} \quad Xiaochen Lian\textsuperscript{2} \quad Yichun Shi\textsuperscript{2} \\ Peihao Zhu\textsuperscript{2} \quad Chen Sun\textsuperscript{1} \quad Peng Wang\textsuperscript{2} \\
\textsuperscript{1}Brown University \quad \textsuperscript{2}ByteDance Seed \quad  \textsuperscript{3}The University of Tokyo
}

\begin{document}
\maketitle
\begin{abstract}

Recent advances in visual generative models have enabled high-fidelity image editing guided by human instructions. However, these models often struggle with complex instructions involving combinatorial editing operations or inter-step dependencies. This difficulty stems from the limitations of two canonical paradigms: (1) single-turn editing, which attempts to apply all instructed edits in one pass, often fails to parse the complex instruction accurately and causes undesired edits; and (2) sequential editing can decompose the task into simpler steps but suffers from compounding errors introduced by the sequential execution, leading to low-fidelity results. To derive a robust solution for complex image editing, we examine editing behaviors of different paradigms under a unified in-context editing framework, and study how the benefits of sequential decomposition can be balanced against its error-accumulation drawbacks. We further develop a synthetic data pipeline that constructs editing tasks of varying instruction complexity, allowing us to curate a large-scale editing dataset with high-quality decomposed sequences. By finetuning on synthetic data, we discovered that with properly designed editing paradigms, sequential decomposition yields robust improvements even as task complexity increases. Furthermore, the decomposition skills learned from synthetic tasks can transfer to real images by co-training with real-world editing data, demonstrating the promise of sim-to-real generalization for tackling complex image editing across broader domains.

\end{abstract}
    
\section{Introduction}

\label{sec:intro}
Image generative models have achieved remarkable progress in synthesizing high-quality visual content aligned with user instructions, unlocking a wide range of applications in text-to-image (T2I) generation~\cite{pmlr-v139-ramesh21a, pmlr-v235-esser24a, BetkerImprovingIG, pmlr-v162-nichol22a, Ramesh2022HierarchicalTI, Saharia2022PhotorealisticTD} and user instruction-guided image editing~\cite{Brooks2022InstructPix2PixLT, Zhao2024UltraEditIF, Wei2024OmniEditBI}. While state-of-the-art T2I models can generate visually compelling images from complex instructions, their counterparts for image editing are often limited to handling a \textit{single type of edit or restricted target regions}, standing in stark contrast to real-world scenarios that usually demand more compositional modifications.

\begin{figure}[t]
    \centering
    \includegraphics[width=\linewidth]{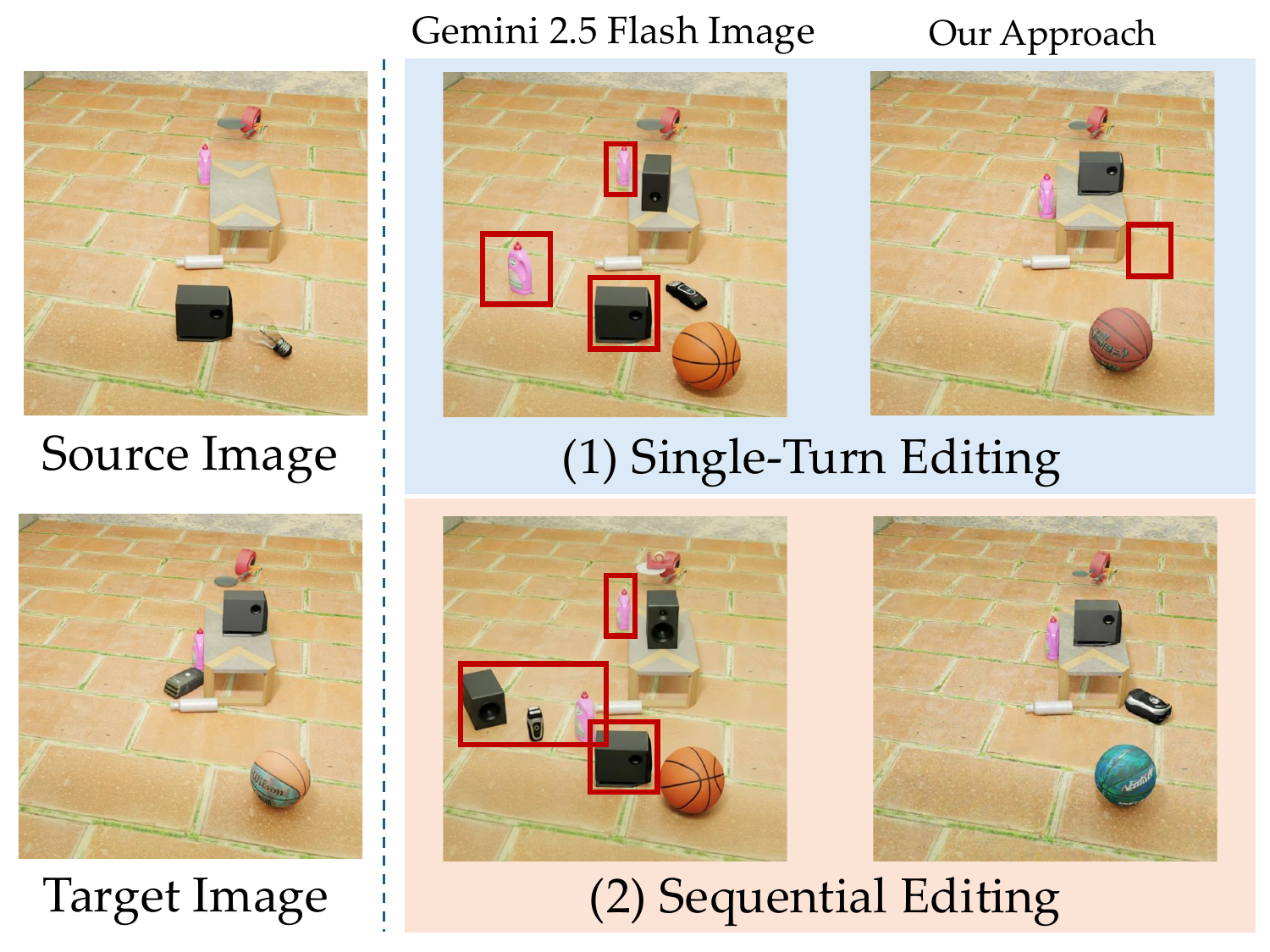}
    \caption{\textbf{Single-Turn Editing versus Sequential Editing} with a complex instruction: "Move the speaker onto the modern coffee table, move the all-purpose cleaner forward, then add a shaver on the floor near the speaker, and replace the lightbulb with a basketball". Incorrect edits are labeled with red boxes on the image.}
    \label{fig:motivation}
    \vspace{-1.4em}
\end{figure}

To enable complex image editing from instructions containing combinatorial editing operations, a model must first identify distinct edits specified by the instruction and then apply them to their corresponding target regions in the source image. Two straightforward paradigms can be considered for this task: (1) single-turn editing, which performs all required edits in a single step, and (2) sequential editing, which decomposes the complex instruction into simpler sub-edits and executes them sequentially. However, neither paradigm reliably produces accurate results under complex instructions. Single-turn editing often yields incomplete or even wrong edits because the model must simultaneously recognize and execute multiple operations from a compounded instruction. Conversely, although sequential editing intuitively simplifies the task through decomposition, it in practice suffers from severe error accumulation~\cite{Yang2025ComplexEditCI, Zhang2023MagicBrushAM, Zhou2025MultiturnCI, Qu2025VINCIEUI, Sheynin2023EmuEP}, where inaccuracies introduced in early steps propagate through subsequent ones, ultimately degrading the fidelity of the final result. As shown in Figure~\ref{fig:motivation}, Gemini 2.5 Flash Image~\cite{gemini2025} fails to perform correct edits in either single-turn or sequential editing, with sequential editing yielding more undesired modifications.

In this work, we aim to develop a robust solution that addresses complex image editing via sequential decomposition. The core challenge lies in achieving an effective trade-off between the advantages of multi-step instruction decomposition and its accompanying issues of error accumulation. To examine this trade-off, we study sequential editing through the lens of in-context editing framework~\cite{openai2025addendum, Qu2025VINCIEUI}, where a multimodal model generates interleaved sequences of decomposed instructions and their corresponding edited images while conditioning on the full history of contextual inputs. This formulation unifies different editing paradigms within a single model and enables a fair comparison of their behaviors. More importantly, by conditioning each edit on rich contextual history, in-context editing makes sequential decompositions more informed and adaptive to intermediate outcomes, while opening up more editing paradigms to explore. These properties make in-context editing a compelling framework for investigating and improving sequential decomposition in complex image editing scenarios.

\begin{figure*}[t]
    \centering
    \includegraphics[width=\textwidth]{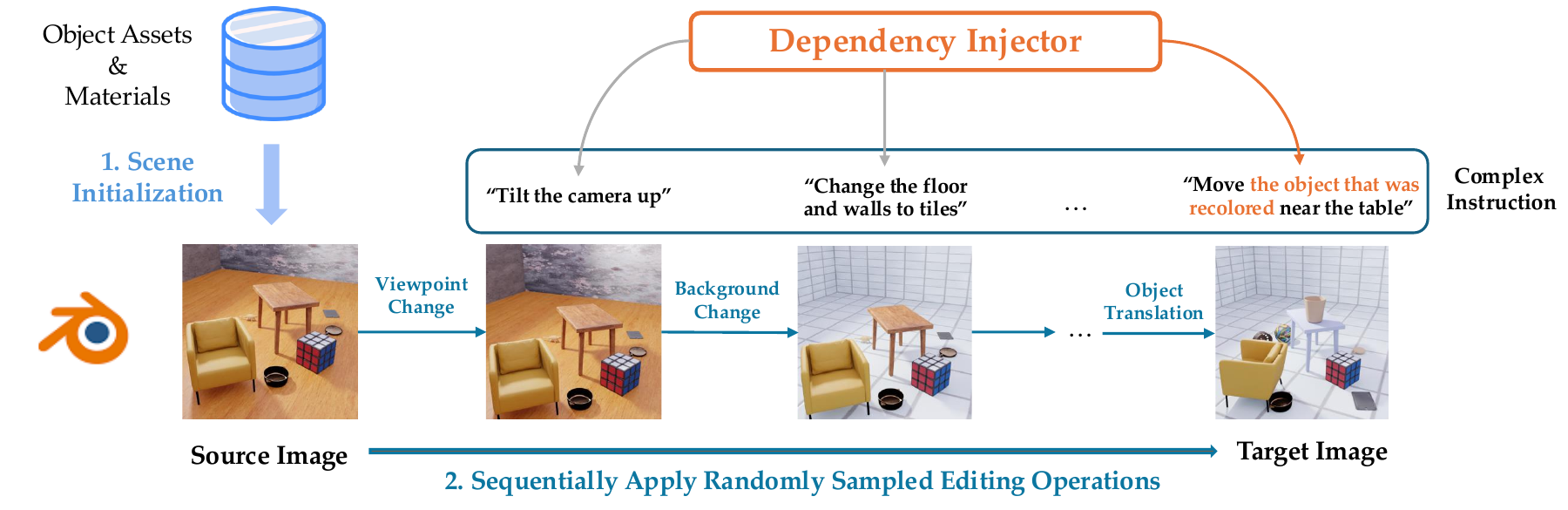}
    \caption{\textbf{Overview of Synthetic Data Pipeline}. We build the synthetic data pipeline on Blender, in which we construct complex editing tasks by sequentially applying editing operations on a randomly initialized scene. After constructing the editing chain, we take the initial and final rendering of the scene as the editing pair and concatenate all operation descriptions as the corresponding complex instruction.}
    \label{fig:data_pipeline}
    \vspace{-1em}
\end{figure*}

To profile editing paradigms under different task complexities, we develop a synthetic data generation pipeline in Blender, where editing tasks are curated by manipulating procedurally generated scenes with a set of atomic editing operations. The pipeline sequentially performs a varying number of editing operations to generate editing sequences, constructing complex tasks by treating the initial scene rendering as the source image and the concatenation of editing operations as the corresponding instruction. Since each operation deterministically modifies the scene, the pipeline can  produce sequences of “decomposed” steps with consistently accurate edits, along with a reference target image rendered from the final scene. To further enhance task complexity, we introduce inter-step dependencies, in which the instruction of a later operation requires referencing results from earlier edits (as illustrated in Figure~\ref{fig:data_pipeline}). This pipeline not only provides a controlled environment for analyzing editing behaviors, where we can modulate task complexity by adjusting the editing length and the dependency ratio, but also enables scalable curation of high-quality editing chains of arbitrary length and dependency without degradation.

We instantiate our in-context editing framework with unified model BAGEL~\cite{deng2025emerging} and finetune it on synthetic editing sequences curated from our pipeline to induce sequential decomposition behaviors. Using the finetuned model, we implement different editing paradigms during inference and evaluate their performance on unseen synthetic tasks of varying complexity. We find that several state-of-the-art editing models suffer performance drops when sequential editing is applied in a zero-shot manner; in contrast, decomposition capabilities learned through our approach not only improve editing performance compared to single-turn editing baselines, but also continue to benefit from additional decomposition steps. Notably, context-guided sequential editing, a paradigm that can flexibly modulate the impact from prior editing results, consistently yields robust improvements across various settings and achieves the best GPT-evaluated scores on tasks with inter-step dependencies. This demonstrates that robust sequential decomposition can be enabled by training on high-quality editing sequences, integrated with a well-designed inference paradigm.  Furthermore, to transfer the decomposition skills to the real-world domain, we finetune BAGEL on a mixture of synthetic editing sequences and real-world single-turn editing pairs. We find that the two-step context-guided editing successfully improves both instruction following and identity preservation on complex real-world editing tasks, highlighting the promise of sim-to-real transfer for enhancing complex image editing across broader domains.

\section{Related Work}
\label{sec:related_work}

\noindent\textbf{In-Context Editing.} 
In-context editing~\cite{openai2025addendum, Qu2025VINCIEUI} built on unified multimodal models~\cite{chen2025unireal, xiao2025omnigen, batifol2025flux, sun2024generative, deng2025emerging} has emerged as a powerful framework that performs edits based on multimodal contexts composed of text and previously generated images. VINCE~\cite{Qu2025VINCIEUI} acquires in-context editing capabilities from large-scale video data, and demonstrates strong multi-turn editing performance with segmentation mask prediction. In our work, we investigate how to derive an effective sequential decomposition strategy within the in-context editing framework to improve editing performance beyond single-turn editing robustly. We base our investigation upon BAGEL~\cite{deng2025emerging}, a unified model with strong performance on existing single-turn editing benchmarks, and study whether and how multi-step sequential decomposition can facilitate complex image editing.

\vspace{2mm}\noindent\textbf{Data Construction for Image Editing.} High-quality data construction is essential to the success of visual generative models. Prior work~\cite{Brooks2022InstructPix2PixLT, Zhao2024UltraEditIF, Sheynin2023EmuEP, Zhang2023MagicBrushAM} has developed automatic pipelines that utilize off-the-shelf image editing models to obtain target images for instruction-based image editing. These approaches are designed for efficient data generation, but are difficult to maintain data quality without carefully designed supervision. We build our data construction pipeline with Blender, which has precise control over the visual changes made by every editing operation and allows us to construct synthetic tasks of varying complexity without quality degradation from sequential modifications.

\vspace{2mm}\noindent\textbf{Editing with Complex Instructions.} Many instruction-based image editing datasets and benchmarks focus on tasks with a single editing type or restricted targeted regions, limiting their applications in real-world scenarios where more complex edits are needed. Recent work~\cite{Yang2025ComplexEditCI} proposes benchmarks with more complicated editing instructions, where each task comprises edits of various types across multiple objects, and attempts to decompose the task via sequential editing. However, even the state-of-the-art editing models~\cite{batifol2025flux, wu2025qwenimagetechnicalreport, gemini2025, openai2025addendum} suffer from severe compounding errors and cause low-fidelity results after sequential execution. %
To overcome the challenge of collecting high-quality decomposed steps on real images for model training across varying task complexities, we acquire decomposition supervision from synthetic tasks and transfer to the real image domain by co-training with real-world editing data.

\section{Synthetic Data Generation}
\label{sec:data_generation}

To systematically investigate the editing behaviors of different paradigms across varying task complexities, we develop a Blender-based synthetic data generation pipeline (as in Figure~\ref{fig:data_pipeline}) that constructs complex editing tasks by sequentially modifying a randomly initialized scene with atomic editing operations. 
In Section~\ref{sec:edit_operation}, we outline atomic editing operations for the editing task curation. In Section~\ref{sec:automatic_pipeline}, we describe the workflow of the automatic pipeline that supports task construction with varied complexity.

\subsection{Editing Operations}
\label{sec:edit_operation}
A diverse set of editing types is essential for constructing complex tasks. We equip our pipeline with 10 different editing operations, which can be divided into three main categories: (1) Local editing focuses on object-level modifications, including addition, removal, replacement, translation, rotation, size change, color change, as well as material change operations; (2) Background change operation replaces the textures of the structural planes (e.g. the floor and walls) to alter the visual appearances of the scene background; (3) Viewpoint change operation adjusts the camera positions by translating (forward or backward), tilting (up or down), panning (left or right) and yawing (left or right). 

Each editing operation can produce distinct visual outcomes depending on its specific parameters, which are randomly sampled during the data generation process to enhance data diversity. Furthermore, we provide language descriptions for each operation by completing the predefined templates with sampled parameters. More details regarding editing operation design can be found in Appendix~\ref{sec:synthetic_data_gen_pipeline}.

\subsection{Automatic Data Pipeline}
\label{sec:automatic_pipeline}
Many existing editing datasets are curated using pretrained generative models, whose quality can be upper-bounded by the failure modes of these models. To circumvent this, we build our pipeline on Blender, which guarantees authentic edits by predefined operations and image quality through advanced photorealistic rendering. The workflow of the data generation pipeline is outlined below.

\vspace{2mm}\noindent\textbf{Scene Initialization.} Given a set of 3D object assets and material textures, we procedurally initialize a scene by constructing an empty room inside Blender, with random textures applied to the floor and wall planes. We then sample up to $N$ different objects and place them on the room floor in a non-collision layout guaranteed by running physics simulations. This creates an initial scene state whose visual complexity depends on object diversity $N$ and their spatial relationships. The rendering of the initial scene will be collected as the source image $I_{0}$ for editing tasks.

\vspace{2mm}\noindent\textbf{Editing Chain Composition.} To construct a complex editing task, we sample an $L$-length chain of editing operations, each paired with a language description $\{T_i\}_{i=1}^{L}$. We then sequentially apply those operations to the initialized scene, and treat the final rendered image as the target of the editing task $I_{L}$. Meanwhile, as we step through all transitions in the editing chain, we can collect these intermediate renderings $\{I_{i}\}_{i=1}^L$ as decomposed editing steps with per-step instruction labels. Finally, we concatenate the descriptions of the editing operations to form a complex instruction $T_c=Concat(\{T_i\}_{i=1}^{L})$. Each generated example comprises an editing triplet $(I_{0}, T_c, I_{L})$ along with the decomposed steps $\{(T_i, I_i)\}_{i=1}^L$. Our method performs each editing operation deterministically, ensuring authentic scene modifications and consistently high-quality decomposition annotations regardless of editing length.

\vspace{2mm}\noindent\textbf{Inter-Step Dependency Injection.} 
When all randomly sampled operations are independent and can be executed in parallel, it reduces task complexity even when the length of the editing chain ($L$) grows. To further enhance task complexity, we inject inter-step dependencies into the task instruction by requiring later operations to reference the results of earlier edits. This design forces the model to first reason over the dependencies to clarify the intended editing operations before executing them. We introduce two types of dependency: (1) operation reference, which selects an object that has been modified in a previous operation and refers to it through its historical edit rather than its explicit name. For example, if a table was rotated in an earlier step, a subsequent instruction may identify it as ``the object that was rotated'' instead of ``the table''. \textit{Notably, we only use reference-based description for objects in overall complex instructions, and reveal their actual labels in decomposed instructions}; and (2) position reference, which specifically applies to object addition and translation operations, is implemented by specifying the target position as the original location of an object before it was moved by an earlier operation. For example, if a chair was moved or removed in an earlier step, a later addition instruction may say, ''Add a book where the chair originally was". When constructing an editing chain with inter-step dependencies, we decide for each operation after the first, independently with probability $p_d$, whether to inject a dependency on an earlier edit.

\section{Method}
\label{sec:method}
We investigate sequential decomposition through in-context editing framework, in which a multimodal model generates interleaved sequences of editing instructions and their corresponding results based on contextual inputs. This framework allows us to implement various editing paradigms within a single model and to study their behaviors from a unified perspective. Moreover, accessing rich historical contexts not only makes the sequential decomposition process more informed, but also enables the exploration of a broader set of sequential editing paradigms. Given a complex editing instruction $T_{c}$ and a source image $I_0$, our framework sequentially generates interleaved editing sequences in the following format: $<T_c, I_0, T_1, I_1, ..., T_L, I_L>$, in which $\{T_i\}_{i=1}^{L}$ are decomposed instructions, $\{I_i\}_{i=1}^{L-1}$ denote intermediate editing outcomes with $I_L$ being the final result for the task. %
\subsection{Model Finetuning}
\label{sec:model_finetuning}
Although many multimodal models are capable of generating interleaved sequences, they usually do not exhibit specialized sequential decomposition behaviors by default due to their general pretraining objectives. To enable such capabilities, we finetune the model by jointly optimizing the following training objectives: (1) Instruction Decomposition: given the complex instruction $T_c$, the source image $I_0$, and all intermediate images $\{I_i\}_{i=1}^{L-1}$, the model is trained to autoregressively predict decomposed instructions between the consecutive images according to the following objective:
\begin{equation}
    L_{D} = -\sum_i\sum_j\log{p_{\theta}(T_{i,j} \mid T_{i, <j}, T_{<i}, I_{<i}, T_c)} \label{eq:decomposition}
\end{equation}
, in which $j$ denotes the token index in per-step instruction $T_{i}$. (2) In-Context Editing: the model takes as input the source image $I_{0}$, and all decomposed text-image pairs $\{(I_i, T_i)\}_{i=1}^L$, in which subsequent image or text tokens may attend to the preceding ones to comprehend context information, and is trained with  rectified flow objective on the intermediate and target images $\{I_i\}_{i=1}^{L}$:
\begin{equation}
    L_{ICE} = \mathbb{E}_{t, X_0, X_1}[||(X_1 - X_0) - v_{\theta}(X_t, t, c)||^2] \label{eq:icd}
\end{equation}
. Assuming we optimize the above objective over the VAE latent space, $X_1$ represents the clean VAE tokens from $\{I_i\}_{i=1}^{L}$, $X_t = tX_1 + (1-t)X_0$ is the noisified latent constructed through linear interpolation between clean VAE tokens and Gaussian noises $X_0 \sim \mathcal{N}(0, I)$, $v_{\theta}$ is the velocity prediction network, and $c$ is the multimodal context, i.e. $c=\{I_{<i}, T_{\leq i}\}$ for $I_i$. Note that we do not include the overall complex instruction $T_c$ in the context $c$ for this training task. This decoupled design allows the generation module to edit images based solely on decomposed instructions without interference from the complex instruction. (3) Single-Turn Editing: the model follows the same objective in Eq.~\ref{eq:icd}, with $c$ being the tuple of the source image and the overall complex instruction $(I_0, T_c)$. This objective retains single-turn editing capabilities by asking the model to edit from $I_0$ to $I_L$ in a single step.

\subsection{Versatile Editing Paradigms}
\label{sec:editing_paradigms}
To support customizable decomposition granularity during inference, we regroup training sequence $\{(T_i, I_i)\} _{i=1}^L$ into $K$ chunks ($1 \leq K \leq L$) randomly, in which each chunk contains a concatenation of $l$ decomposed instructions $\{T_j, ..., T_{j+l-1}\}$ and their editing result $I_{j+l-1}$, and omits $\{I_{j+k}\}_{k=0}^{l-2}$. We then finetune the model to automatically decompose the task into varying numbers of chunks as specified by the value of $K$ appended to the complex instruction $T_c$. Specifically, $K=1$ reduces the objective to single-turn editing as all intermediate images will be omitted; by setting $K=L$, the model is finetuned over original decomposed sequences. Furthermore, when optimizing in-context editing, we randomly drop text and clean image tokens from the context to enable classifier-free guidance for interleaved generation, which encompasses versatile editing paradigms during inference. We outline three different editing paradigms enabled by our unified framework below.

\vspace{2mm}\noindent\textbf{Single-Turn Editing.} As described above, we can implement single-turn editing by appending $K = 1$ to the overall complex instruction. The model will first generate a compositional instruction $\hat{T}_c$ as the concatenation of all decomposed operations, and then directly edit the source image $I_0$ following $\hat{T}_c$. When handling instructions with operation reference, the model will translate the reference-based descriptions and reveal the explicit object labels in $\hat{T}_c$.

\vspace{2mm}\noindent\textbf{Full-Context Sequential Editing (FCSE).} When $K > 1$, the model decomposes the instruction into $K$ groups of edits, each followed by the generation of their corresponding editing result. When generating $I_i$, we fix the historical editing sequence $\{T_{<i}, I_{<i-1}\}$ as input, and compute guidance with $T_{i}$ and $I_{i-1}$ respectively to emphasize the impact from the current decomposed instruction as well as the last editing result. We term this paradigm as Full-Context Sequential Editing, and compute CFG as follows: $v_{\text{FCSE}} = v_{\theta}(X_t, t, \{T_{<i}, I_{<i-1}\}) + \gamma_I(v_{\theta} (X_t, t, \{T_{<i}, I_{<i}\}) - v_{\theta}(\\X_t, t, \{T_{<i}, I_{<i-1}\})) + \gamma_T(v_{\theta} (X_t, t, \{T_{\leq i}, I_{<i}\}) - v_{\theta} (X_t, t, \{T_{<i}, I_{<i}\}))$

\vspace{2mm}\noindent\textbf{Context-Guided Sequential Editing (CGSE).} In full-context sequential editing, the error introduced by previous editing results might accumulate through sequential execution, leading to degraded performance. We can mitigate this by isolating prior editing results as a separate guidance term whose strength can be modulated by a coefficient $\gamma_{ctx}$. For every step $i$, we start with the velocity $v_{d}$ that directly edits $I_0$ to $I_i$ by computing CFG with source image $I_{0}$ and the concatenation of the decomposed instructions $\{T_j\}_{j=1}^i$: $v_{d} = v_{\theta} (X_t, t) + \gamma_I(v_{\theta} (X_t, t, I_0) - v_{\theta} (X_t, t)) + \gamma_T(v_{\theta} (X_t, t, I_0, \{T_j\}_{j=1}^i) - v_{\theta} (X_t, t, I_0))$. Then, we apply the guidance from contextual editing results (termed "context guidance") on top of $v_{d}$, with a strength coefficient $\gamma_{ctx}$ regulating its effect on the current edit. Moreover, we can arbitrarily select historical edits to compose a guidance context: $v_{\text{CGSE}} = v_{d} + \gamma_{ctx}(v_{\theta} (X_t, t, I_0, \{T_j\}_{j=1}^i, \{I_j\}_{j=m}^n) - v_{d})$, where $m \leq n < i$. For example, we can include all previous editing results $\{I_j\}_{j=1}^{i-1}$ for context guidance computation; alternatively, we can use the last image $I_{i-1}$ alone to highlight its impact on the result. Compared to full-context sequential editing, this paradigm provides more control over the influence from history edits through $\gamma_{ctx}$ and the guidance context composition, allowing us to balance the auxiliary effect against the error introduced by the task decomposition. More details can be found in Table~\ref{tab:cfg_comparison}.

\subsection{Co-training with Real-World Editing Data}
\label{sec:cotraining_with_real}
While the synthetic data pipeline enables scalable task curation with varying complexity, finetuning models solely on synthetic data limits their applicability to broader domains. To transfer the decomposition skills learned from synthetic tasks to real-world scenarios, we finetune our model on a mixture of synthetic editing sequences with decomposition annotations and single-turn editing data on real images. This allow us to effectively combines the large-scale and high-quality data from both domains, allowing the model to transfer decomposition abilities obtained from synthetic tasks while retaining editing capability on real images.

\section{Experiment}
\label{sec:experiment}

In this section, we first evaluate different editing paradigms under various task complexities in the synthetic domain and investigate how decomposition should be properly utilized to facilitate image editing tasks as complexity increases. Then, we study whether the decomposition skills learned from synthetic tasks can generalize to real-world editing tasks by co-training with real editing data.

\subsection{Experimental Setup} 

\noindent\textbf{Synthetic Data Generation.} We use 2,246 3D assets from Digital Twin Catalog~\cite{Dong_2025_CVPR} and PolyHaven~\cite{polyhaven} for easy access to their semantic labels, and 1,866 materials from CC0 Textures~\cite{cctextures} as basic elements for scene construction. Under our pipeline, we generate a large-scale editing dataset that %
includes 60K editing sequences by naively chaining random operations and 34K sequences with historical dependency injected. 
We initialize every scene with up to $N=9$ objects and vary $L$ from 3 to 17 to construct editing chains with a varying number of operations, and set $p_d = 0.5$ when inter-step dependencies are enabled. Detailed dataset analysis is provided in Appendix~\ref{sec:dataset_analysis}.

\vspace{2mm}\noindent\textbf{Implementation Details.} We instantiate our in-context editing framework with BAGEL~\cite{deng2025emerging}, a unified multimodal model that natively supports interleaved generation of multimodal sequences, and finetune it on curated synthetic data with the hyperparameters provided in Appendix~\ref{sec:exp_details}. Then we implement different editing paradigms at inference time as described in Section~\ref{sec:editing_paradigms} for evaluation. To mitigate compounding errors in context-guided sequential editing, we take the editing result from the previous step alone to compute context guidance.
To transfer decomposition skills to the real-world domain, we separately finetune BAGEL on a mixture of the curated synthetic data and the single-turn editing data from Pico-Banana~\cite{qian2025picobanana400klargescaledatasettextguided}. As BAGEL is pre-trained on large-scale real-world datasets, we optimize the model for 1,000 steps to investigate whether our framework enables efficient sim-to-real generalization.

\vspace{2mm}\noindent\textbf{Benchmarks.} To evaluate different editing paradigms on synthetic tasks of varying complexity, we generate 500 editing sequences using \textit{unseen} objects for two settings: (1) naive operation chains without inter-step dependencies (Independent Chain), and (2) operation chains augmented with inter-step dependencies (Dependent Chain), in which each instruction has at least one operation that require referencing a historical edit. The number of editing operations and the ratio of dependent operations are evenly distributed in each setting when applicable. To evaluate sim-to-real generalization, we use Complex-Edit~\cite{Yang2025ComplexEditCI}, a benchmark constructed to evaluate editing performance with complex instructions. We take 531 tasks from their real image editing split and supply the model with the concatenation of 8 atomic instructions from each task for evaluation.

\begin{table}[t]
\centering
\caption{\textbf{Evaluations on tasks without inter-step dependencies.} ``FCSE'' denotes ``Full-Context Sequential Editing'' and ``CGSE'' denotes ``Context-Guided Sequential Editing'' with $\gamma_{ctx} = 2.5$.}
\label{tab:independent_chain_eval}
\resizebox{\columnwidth}{!}{%
\begin{tabular}{@{}l|lll|lll@{}}
\toprule
\multirow{2}{*}{Independent Chain}           & \multicolumn{3}{c|}{Single-Turn Editing} & \multicolumn{3}{c}{Sequential Editing} \\ \cmidrule(l){2-7} 
                                    & DINO-I $\uparrow$ & DINO-D $\uparrow$ & GPT-5 $\uparrow$ & DINO-I $\uparrow$ & DINO-D $\uparrow$ & GPT-5 $\uparrow$ \\ \midrule
Qwen-Image                          & 0.614  & 0.372  & 2.27  & -      & -      & -     \\
GPT-4o*                             & 0.564  & 0.411  & \textbf{5.35}  & 0.506  & 0.393  & \textbf{4.61}  \\
Gemini 2.5 Flash Image              & 0.693  & 0.425  & 3.88  & 0.676  & 0.446  & 4.09  \\
BAGEL  (Zero-Shot)                  & 0.541  & 0.385  & 2.43  & 0.500  & 0.372  & 2.24  \\ \midrule
BAGEL (Finetuned)                   & \textbf{0.756}  & \textbf{0.576}  & \textbf{4.0}   & -      & -      & -     \\ 
~~w/ FCSE  ($K = 3$) & -         & -        & -        & 0.738       & 0.557       & 4.14       \\
~~w/ FCSE  ($K = 5$) & -         & -        & -        & 0.711       & 0.529       & 4.16       \\
~~w/ CGSE  ($K = 3$) & -      & -      & -     & \textbf{0.749}  & \textbf{0.562}  & 4.15  \\
~~w/ CGSE  ($K = 5$) & -      & -      & -     & 0.737  & 0.549  & \textbf{4.20}  \\ \bottomrule
\end{tabular}%
}
\vspace{-1em}
\end{table}

\vspace{2mm}\noindent\textbf{Metrics.} To evaluate editing performance on synthetic tasks, we follow prior work~\cite{Chang2025ByteMorphBI, Chen2023AnyDoorZO, Zhang2023MagicBrushAM} to adopt two similarity metrics based on DINOv3~\cite{Simeoni2025DINOv3} features, DINO-I and DINO-D, to measure the visual discrepancy between generated images and reference target images. We compute DINO-D as $cos(\boldsymbol{C}(I_{gen}) - \boldsymbol{C}(I_{src}), \boldsymbol{C}(I_{tgt}) - \boldsymbol{C}(I_{src}))$ to measure similarity over editing directions, where $\boldsymbol{C}$ denotes DINOv3 encoder, and $I_{src}, I_{tgt}, I_{gen}$ denote the source image, the target image and the generated image respectively. 
However, since most editing operations modify only one object that occupies a local region of the entire scene and cause small pixel-wise changes, high similarity scores can be achieved when the evaluated images share similar backgrounds.
To address this, we also employ GPT-5 to evaluate whether the edited image accurately reflects all changes requested by the complex instruction without introducing unintended edits. For each result, GPT-5 assigns a score from 0 to 10, with 10 indicating a perfect edit. 
In sim-to-real evaluations, we adopt the VLM-based metrics from Complex-Edit, which measure the quality of editing results from three dimensions: Instruction Following (IF), Identity Preservation (IP), and Perceptual Quality (PQ). %

\begin{table}[t]
\centering
\caption{\textbf{Evaluations on tasks with inter-step dependencies.} ``FCSE'' denotes ``Full-Context Sequential Editing'' and ``CGSE'' denotes ``Context-Guided Sequential Editing'' with $\gamma_{ctx} = 2.5$.}
\label{tab:dependent_chain_eval}
\resizebox{\columnwidth}{!}{%
\begin{tabular}{@{}l|lll|lll@{}}
\toprule
\multirow{2}{*}{Dependent Chain}             & \multicolumn{3}{c|}{Single-Turn Editing} & \multicolumn{3}{c}{Sequential Editing} \\ \cmidrule(l){2-7} 
                                    & DINO-I $\uparrow$     & DINO-D $\uparrow$     & GPT-5 $\uparrow$    & DINO-I $\uparrow$ & DINO-D $\uparrow$ & GPT-5 $\uparrow$ \\ \midrule
Qwen-Image                          & 0.632       & 0.358       & 2.02      & -      & -      & -     \\
GPT-4o*                             & 0.579       & 0.388       & \textbf{3.77}      & 0.507  & 0.370  & 3.09  \\
Gemini 2.5 Flash Image              & 0.712       & 0.408       & 3.23      & 0.698  & 0.419  & 3.28  \\
BAGEL (Zero-Shot)                   & 0.579       & 0.372       & 2.17      & 0.552  & 0.382  & 2.03  \\ \midrule
BAGEL (Finetuned)                   & 0.776/\textbf{0.791} & 0.553/\textbf{0.578} & 3.57/\textbf{3.91} & -      & -      & -     \\ 
~~w/ FCSE ($K = 3$) & -         & -        & -        & 0.756       & 0.536       & 4.06       \\
~~w/ FCSE ($K = 5$) & -         & -        & -        & 0.723       & 0.509       & \textbf{4.13}       \\
~~w/ CGSE ($K = 3$) & -           & -           & -         & \textbf{0.779}  & \textbf{0.555}  & 4.04  \\
~~w/ CGSE ($K = 5$) & -           & -           & -         & 0.762  & 0.533  & \textbf{4.14}  \\ \bottomrule
\end{tabular}%
}
\vspace{-1em}
\end{table}

\subsection{Complex Image Editing in Synthetic Domain}

\begin{figure*}[t]
    \centering
    \includegraphics[width=\linewidth]{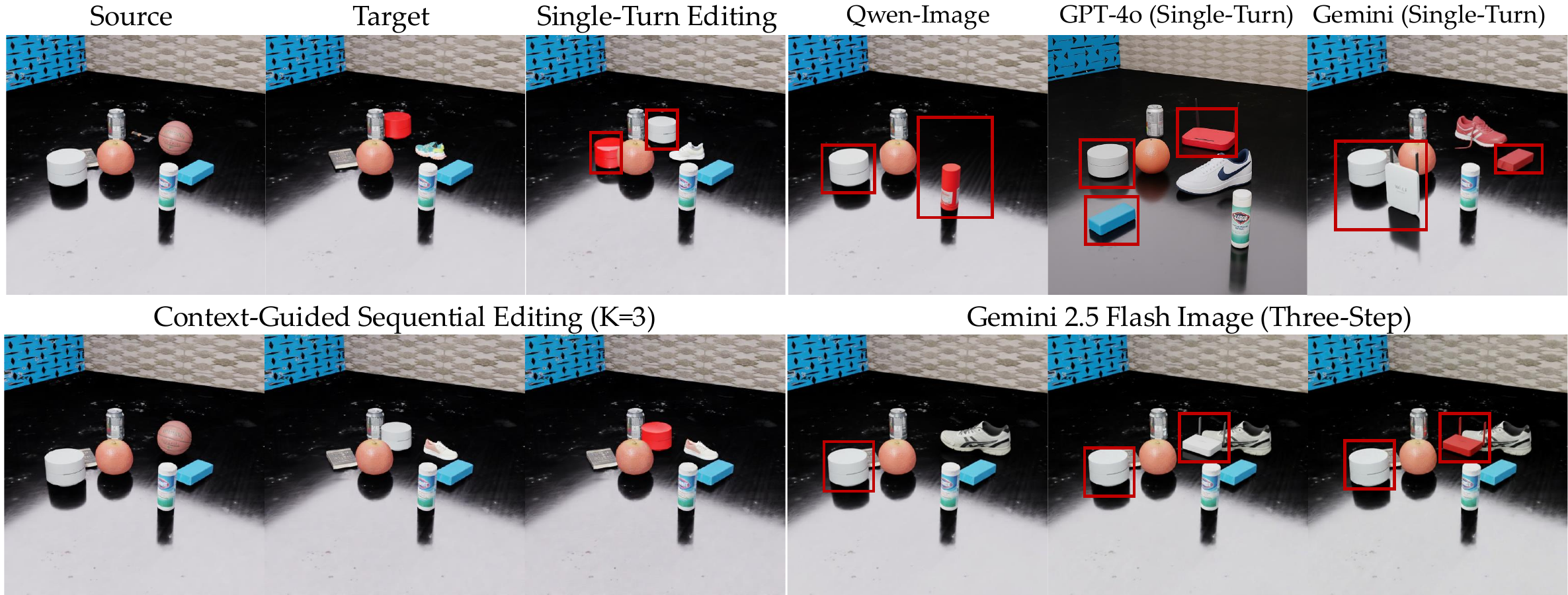}
    \caption{Qualitative comparisons on a synthetic task with inter-step dependency. Instruction: ``Remove the hatchet, replace the basketball with a shoe, relocate the wifi router where the hatchet initially was, and paint the object that was moved red''. ``Single-Turn Editing'' refers to the single-turn editing result from finetuned BAGEL. Red boxes indicate incorrect edits. More examples can be found in Appendix~\ref{sec:more_qualitative}.}
    \label{fig:synthetic_dependent_result}
    \vspace{-1em}
\end{figure*}

In Table~\ref{tab:independent_chain_eval}, we compare single-turn editing against sequential editing paradigms on 500 \textit{unseen} editing tasks without inter-step dependencies, and report the zero-shot editing performance for four editing model baselines, Qwen-Image~\cite{Wu2025QwenImageTR}, GPT-4o~\cite{openai2025addendum}, Gemini 2.5 Flash Image~\cite{gemini2025}, and BAGEL (Zero-Shot)~\cite{deng2025emerging}. In a sequential editing setup, we decompose the instruction by default into 3 steps for baseline models. We observe that both BAGEL (Zero-Shot) and GPT-4o suffer from severe performance degradation during sequential editing, underscoring the challenge of reliably improving performance through naive task decomposition.
After finetuning on synthetic data, we find that BAGEL achieves strong single-turn editing performance that surpasses all baselines except for GPT-4o. 
Meanwhile, effective improvements are achieved by both full-context sequential editing and context-guided editing paradigms, as indicated by the GPT-based metric. 
We also note that nearly all methods exhibit a decline in similarity metrics after applying sequential editing. We attribute this to the sensitivity of pretrained visual features to the inevitable artifacts introduced during sequential editing, where even minor artifacts that do not affect the semantic correctness of the edit can compromise feature similarity. While both sequential editing paradigms can benefit from increased decomposition steps, CGSE remains robust on similarity metrics, highlighting its resilience to potential compounding errors.

Table~\ref{tab:dependent_chain_eval} shows the editing performance on tasks where inter-step dependencies are introduced to enhance complexity. In this setting, effective decomposition requires more than simply dissecting the instruction; it also demands translating the embedded dependencies to clarify the intended editing operations. We discovered that all baseline methods achieve lower performance on tasks with inter-step dependencies, which reflects the genuine difficulty introduced by inter-step dependencies. Additionally, we report two single-turn editing performances for finetuned BAGEL, in which the first number corresponds to the performance where the model directly edits the image based on the complex instruction with operation references, and the second number shows the performance using the instruction clarified by the model as described in Section~\ref{sec:editing_paradigms}. We observe that our finetuned model achieves the best performance on both single-turn and sequential editing setups when the dependencies are clarified. Notably, 3-step context-guided editing outperforms all sequential editing methods on similarity metrics, and 5-step context-guided editing obtains the best GPT-evaluated score. Figure~\ref{fig:synthetic_dependent_result} illustrates that context-guided editing successfully resolves the task with inter-step dependencies through sequential decomposition, whereas single-turn editing and other baselines fail. This again demonstrates that the benefits of sequential decomposition can outweigh its drawbacks from error accumulation, enabling robust improvement on complex image editing.

\vspace{2mm}\noindent\textbf{Task Complexity Analysis.} %
The number of editing operations encoded in the instruction is a primary factor contributing to task complexity. In Table~\ref{tab:ablation_num_edits}, we group tasks into 4 bins based on the number of editing operations and report the aggregated GPT-based scores of each editing paradigm. We show that all editing paradigms encounter a performance drop when the number of editing operations increases. However, sequential decomposition consistently outperforms single-turn editing even when the length of the editing chain grows. Both sequential editing paradigms achieve their best performance with 5 decomposed steps. For each task, we also consider the ratio of editing operations that are dependent on previous edits as another complexity dimension. In Table~\ref{tab:ablation_dependency_ratio}, we compare editing paradigms under different ranges of task dependency ratio, and observe that the gains from sequential decomposition remain robust across all settings. Specifically, the 5-step full-context sequential editing and context-guided editing paradigms perform best in the [0, 0.2] and [0.2, 0.4] ranges, respectively. We note that an excessively high dependency ratio (\eg, $> 0.6$) may include short chains with low task complexity, where high performance is easier to achieve.

\begin{table}[t]
\caption{\textbf{Analysis with different numbers of editing operations}}
\vspace{-0.5em}
\label{tab:ablation_num_edits}
\resizebox{\columnwidth}{!}{%
\begin{tabular}{@{}lllll@{}}
\toprule
\# Editing Operations         & 3-5       & 6-9       & 10-13     & \textgreater{}13 \\ \midrule
Single-Turn Editing           & 4.18/4.28 & 3.61/4.19 & 3.43/3.68 & 3.19/3.56        \\
FCSE ($K=3$) & 4.67      & 4.27      & 3.84      & 3.58             \\
FCSE ($K=5$) & 4.56      & \textbf{4.38}      & 3.97      & \textbf{3.70}             \\
CGSE ($K=3$)    & 4.75      & 4.19      & 3.83      & 3.56             \\
CGSE ($K=5$)    & \textbf{4.81}      & 4.29      & \textbf{4.02}      & 3.58             \\ \bottomrule
\end{tabular}%

}
\vspace{-1.1em}
\end{table}

\begin{table}[t]
\caption{\textbf{Analysis with different dependency ratios}}
\vspace{-0.5em}
\label{tab:ablation_dependency_ratio}
\resizebox{\columnwidth}{!}{%
\begin{tabular}{@{}lllll@{}}
\toprule
Dependency Ratio           & 0-0.2     & 0.2-0.4   & 0.4-0.6   & \textgreater{}0.6 \\ \midrule
Single-Turn Editing           & 3.92/3.94 & 3.62/3.93 & 3.22/3.71 & 3.29/4.46         \\
FCSE ($K=3$) & 4.05      & 4.02      & 3.96      & \textbf{4.78}              \\
FCSE ($K=5$) & \textbf{4.34}      & 4.04      & \textbf{4.13}      & 3.96              \\
CGSE ($K=3$)    & 4.16      & 3.97      & 4.08      & 4.07              \\
CGSE ($K=5$)    & 4.22      & \textbf{4.23}      & 3.75      & 4.75              \\ \bottomrule
\end{tabular}%
}
\vspace{-1.8em}
\end{table}

\begin{figure}[t]
    \centering
    \includegraphics[width=\linewidth]{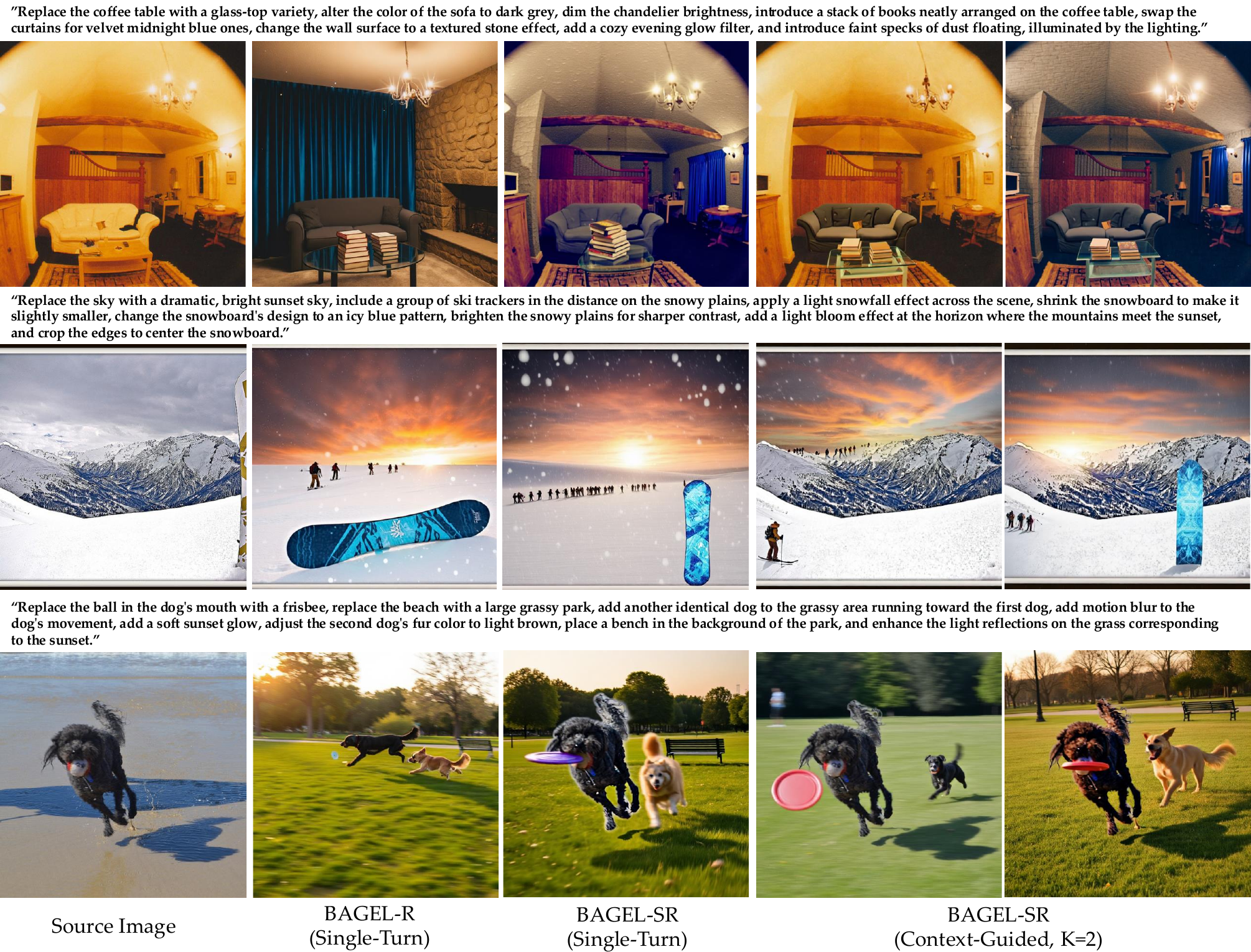}
    \caption{Qualitative comparisons on real-world editing tasks. BAGEL-SR (CGSE, K=2) exhibits stronger detail-preserving abilities compared to the baselines. See more results in Appendix~\ref{sec:more_qualitative}.}
    \vspace{-1.0em}
    \label{fig:real_qualitative_result}
\end{figure}

\subsection{Sim-to-Real Generalization}
To study whether the sequential decomposition skills learned from the synthetic domain can efficiently generalize to real-world editing, we finetune BAGEL on a mixture of synthetic editing sequences and single-turn real-world editing data for 1,000 steps. %
In Table~\ref{tab:sim_to_real}, we report the performance for four models under different editing paradigms on Complex-Edit~\cite{Yang2025ComplexEditCI}: (1) BAGEL-Zero-Shot, (2) BAGEL-S, where only synthetic editing sequences are used for finetuning, (3) BAGEL-R, in which only single-turn real editing data is used for finetuning, and (4) BAGEL-SR, where we finetune BAGEL on both synthetic and real editing data.

The results from BAGEL-Zero-Shot and BAGEL-R show that naively decomposing the task into multiple steps leads to performance degradation across all metrics, and finetuning on synthetic data alone does not naturally enable generalization. However, we find that when co-training with real editing data, synthetic data enables effective improvements over identity preservation, with context-guided editing paradigms consistently outperforming single-turn baselines. In Figure~\ref{fig:real_qualitative_result}, we observe that BAGEL-SR with context-guided editing indeed exhibits stronger detail-preserving capabilities. We hypothesize that synthetic editing sequences provide dense supervision signals that preserve unedited scene elements during training, and such capabilities can be effectively transferred to real-world editing tasks. Moreover, we observe that two-step context-guided editing performs the best on instruction following and identity preservation, reaffirming that the decomposition skills learned from synthetic data can be further utilized to improve real-world editing with complex instructions.

\begin{figure}[ht]
    \centering
    \vspace{-0.5em}
    \includegraphics[width=\linewidth]{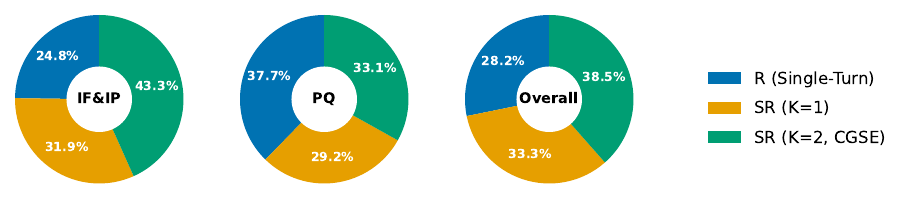}
    \caption{\textbf{Human Preference Study on Complex-Edit.}}
    \vspace{-0.5em}
    \label{fig:user_study}
\end{figure}

Additionally, we perform a human preference study with 33 participants. Each participant was assigned 15 random tests comparing editing results from three finetuned BAGEL models\textemdash R (Single-Turn), SR ($K=1$), and SR ($K=2$, CGSE, $\gamma_{ctx}=0.5$). Each test asks the participant to select the best image based on instruction following \& identity preservation (IF\&IP), and perceptual quality (PQ), respectively, along with the best compromise (Overall). In Figure~\ref{fig:user_study}, we observe that user preference positively correlates with VLM-based evaluation results. Furthermore, although SR (CGSE) is slightly (4.6\%) less preferable than R on PQ, it remains the top-performing approach for both IF\&IP and Overall as recognized by human evaluators.

\begin{table}[t]
\centering
\caption{\textbf{Evaluations on sim-to-real generalization.} Additional baselines are provided in Table~\ref{tab:additional_real_baselines}.}
\label{tab:sim_to_real}
\resizebox{\columnwidth}{!}{%
\begin{tabular}{@{}cllll@{}}
\toprule
Complex-Edit    & Editing Paradigm                   & IF $\uparrow$  & IP $\uparrow$  & PQ $\uparrow$  \\ \midrule

\multirow{2}{*}{Zero-Shot} & Single-Turn Editing              & 7.88 & 5.22 & 5.96 \\
 & 2-Step Seq. Editing  & 7.18 & 4.62 & 5.46 \\ \midrule
\multirow{2}{*}{R} & Single-Turn Editing              & 8.20 & 6.00 & \textbf{7.77} \\
 & 2-Step Seq. Editing & 7.72 & 5.59 & 7.12 \\ \midrule
\multirow{2}{*}{S} & Single-Turn Editing    & 7.14 & 5.79 & 5.77 \\
 & CGSE ($K = 2, \gamma_{ctx} = 0.5$)  & 7.18 & 5.92 & 5.75 \\ \midrule
\multirow{5}{*}{SR} & Single-Turn Editing             & 8.23 & 6.26 & 6.99 \\ 
 & FCSE ($K=2$) & 8.13 & 5.94 & 6.65 \\
 & CGSE ($K = 2, \gamma_{ctx} = 0.5$) & \textbf{8.25} & \textbf{6.38} & 6.96 \\ 
 & CGSE ($K=2, \gamma_{ctx} = 0.7$) & 8.22 & 6.36 & 6.88 \\
 & CGSE ($K = 3, \gamma_{ctx} = 0.5$) & 8.10 & 6.35 & 6.86 \\ \bottomrule
\end{tabular}%
}
\vspace{-1.0em}
\end{table}

Nonetheless, we find that full-context sequential editing does not generalize well to real editing tasks and even underperforms single-turn editing, likely due to its reliance on high-quality contextual information from real images. Unlike in the synthetic domain, naively increasing decomposition steps does not yield clear performance gains. Finally, finetuning on synthetic data may impair the perceptual quality of generated images. While our work primarily focuses on improving semantic consistency between editing processes and the instruction, enhancing the perceptual quality of the sequentially edited images is left for future work.

\section{Conclusion}
\label{sec:conclusion}

In this work, we investigate how to robustly improve complex image editing tasks via sequential decomposition. The central challenge lies in balancing the benefits of multi-step decomposition with the error accumulation introduced by sequential execution. We examine sequential decomposition within an in-context editing framework and develop a synthetic data pipeline that constructs editing tasks of varying instruction complexity. We observe that finetuning models on the curated editing sequences enables sequential decomposition to consistently improve the execution of complex editing instructions, with the strongest gains observed on tasks involving inter-step dependencies. Particularly, context-guided sequential editing demonstrates robust performance across diverse evaluation settings and outperforms other editing paradigms. Finally, we discover that the decomposition skills learned from synthetic tasks can be further transferred to complex real-world editing scenarios by co-training with real single-turn editing data.

\section*{Acknowledgement}
We would like to thank Bo Liu, Qi Zhao, Yuwei Guo, Calvin Luo, and Nate Gillman for their helpful discussion.

\clearpage
\appendix
\setcounter{page}{1}
\setcounter{table}{0}
\renewcommand{\thetable}{A\arabic{table}}
\setcounter{figure}{0}
\renewcommand{\thefigure}{A\arabic{figure}}
\renewcommand\theHtable{Appendix.\thetable}
\maketitlesupplementary

\section{Synthetic Data Generation}

\label{sec:synthetic_data_gen_pipeline}
\subsection{Scene Initialization}
We implement our synthetic data generation pipeline on top of BlenderProc~\cite{Denninger2023}, a Blender-based pipeline for photorealistic image rendering. To create a photorealistic 3D scene, we first construct an empty rectangular room of size 8x8x3, in which we create fundamental structural planes, including a floor, a ceiling, and four walls. These structural planes are placed in a coordinate system with the floor at $z=0$ and walls aligned to the room boundaries. Additionally, a light plane (embedded in the ceiling) is added to ensure basic illumination. Then we randomly sample textures from CC0 Textures for the floor and walls. 

After the base empty room is constructed, we start to populate it with movable objects selected from an asset catalog, which is compiled beforehand to list available objects with their metadata, including semantic labels, categories, and estimated sizes. To ensure every scene has at least one object that can serve as a supporter to other objects, we first sample an object from predefined supporter categories (\eg, tables, desks), and place the object at the center of the room. Before we sample other room objects, we first compute a safe object size range based on the room dimensions, object count, and camera field of view to prevent choosing objects that are too large to fit comfortably in the scene, with which we can filter objects according to their estimated sizes from the catalog. We then sample $N=9$ unique objects in a single batch from the catalog, and place them sequentially into a constrained area where the camera can easily obtain good visibility on all objects. During this process, each placement finds a collision-free position for the object on the floor using collision checkers from BlenderProc. If no collision-free positions can be found for the objects, we skip the placement of the current object and move on to the next one. When the whole batch of selected objects are processed, we run physics simulation for all objects simultaneously to improve scene realism. Once the scene are populated with objects, we sample a camera pose that provides a good coverage of scene content, ensuring the spawned objects are adequately visible and well-framed in the rendered image. By default, we render every scene into a $512 \times 512$ image with 256 samples for each pixel.

\begin{figure*}[t]
    \centering
    \includegraphics[width=\textwidth]{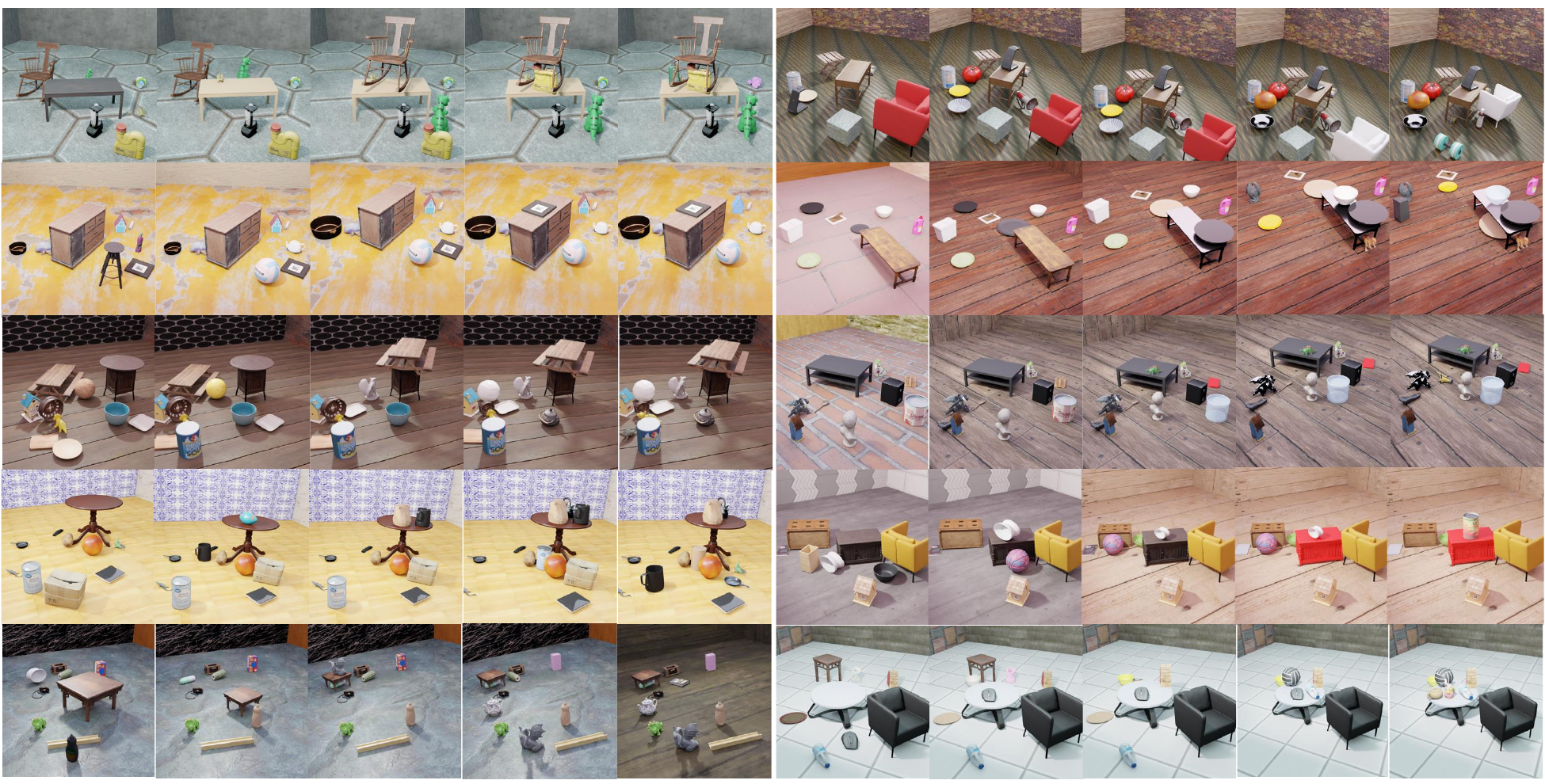}
    \vspace{-2em}
    \caption{\textbf{Visualizations of constructed editing sequence samples}}
    \label{fig:dataset_gallery_main}
\end{figure*}

\begin{figure*}[t]
    \centering
    \includegraphics[width=\textwidth]{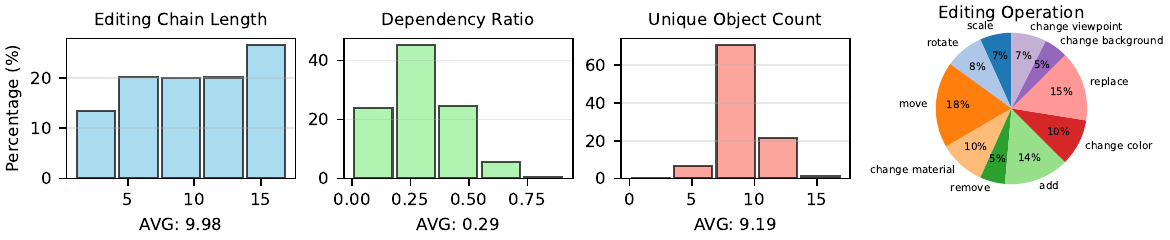}
    \vspace{-2em}
    \caption{\textbf{Statistical analysis of synthetic editing sequences}}
    \label{fig:dataset_analysis}
    \vspace{-1em}
\end{figure*}

Furthermore, we record the metadata (\eg, object labels, categories, poses, sizes, etc.), and the spatial relationships between objects into a registry for all loaded objects in the scene, providing a clearly documented initial state where the subsequent editing process can operate on.
\subsection{Editing Operation Design}
To enable complex edits in our pipeline, we predefined 10 editing operations in total, each associated with a set of parameters. Each operation either performs an edit on a single object in the scene, changes the camera position, or changes the scene background. We outline the predefined editing operations in detail below:

\vspace{1mm}\noindent\textbf{Object Addition} introduces a new object of a novel category under a specified size constraint into the scene to encourage diversity. We support two modes for regular object addition: adding an object on top of a supporter object or near another object on the floor. Additionally, to enable position reference dependency, we allow the operation to directly take in a 3D coordinate of the initial location of an object that has been moved. When adding the object, we also run collision and visibility checking to maintain the visual quality of the rendered image.

\vspace{1mm}\noindent\textbf{Object Removal} deletes an object from the scene. However, we do not allow the structural planes (\eg, floors and walls) or objects that are supporting other objects to be removed.

\vspace{1mm}\noindent\textbf{Object Translation} repositions a non-supporter object while maintaining collision-free placement and camera visibility. We support three regular translation modes, allowing the target object to move by direction (left/right/forward/backward), near a reference object with a distance constraint, or onto another object. Similarly, we allow the operation to take in 3D coordinates directly for position reference.

\vspace{1mm}\noindent\textbf{Object Rotation} rotates the object around its vertical axis, with a degree sampled from predefined candidates (60\textdegree{}, 90\textdegree{}, 120\textdegree{}, 180\textdegree{}) and a direction sampled from ``clockwise'' or ``counter-clockwise''.

\vspace{1mm}\noindent\textbf{Object Replacement} replaces an existing object with a different one sampled from the catalog. The newly placed object inherits the $xy$ location of the original objects. We also run collision checks for the new object and perform slight adjustments if collisions occur.

\vspace{1mm}\noindent\textbf{Color Change} replace the color of an existing object with new one selected from a predefined palette of 12 colors, which includes ``red'', ``blue'', ``green'', ``yellow'', ``orange'', ``purple'', ``pink'', ``white'', ``black'', ``gray'', ``brown'', and ``beige''.

\vspace{1mm}\noindent\textbf{Material Change} replace the material of an existing object with a new one chosen from a predefined set of 5 materials, which includes ``wood'', ``metal'', ``plastic'', ``fabric'', and ``glass''.

\vspace{1mm}\noindent\textbf{Size Change} applies uniform size transformations to an existing object with a scale sampled from predefined candidates (0.6, 0.8, 1.2, 1.5, 1.8) with collision checks.

\vspace{1mm}\noindent\textbf{Viewpoint Change} includes four types of camera pose transformation: (1) translation (0.4m forward or backward); (2) panning (0.4m left or right); (3) tilting (5\textdegree{} up or down); (4) yawing (5\textdegree{} left or right). Meanwhile, the visibility of all objects should remain the same after the camera pose transformation.

\vspace{2mm}\noindent\textbf{Background Change} modifies the visual appearances of structural surfaces (\eg, floors and walls) by replacing their textures with a new material sampled from CC0 Textures. Every background change operation can either change the texture of the floor or both the floor and the walls.

\begin{table*}[t]
\centering
\caption{\textbf{GPT-5 prompts for two-step evaluation}}
\label{tab:eval_prompts}

\resizebox{\textwidth}{!}{
\begin{tabular}{@{}p{0.14\linewidth} p{0.76\linewidth}@{}}
\toprule

System Prompt &
You are an expert evaluator for image editing. You will first be provided with a complex editing instruction and the source image to be edited, please analyze the complex task by rigorously tracking the editing transitions, and list all modifications that need to be reflected in the final edited image. Then you will be given an edited image for evaluation. You will need to examine carefully if the edited image accurately reflects all the required changes according to your reasoning. A successful edit should not have any extra changes that are not required by the instruction. The edited image should have minimal changes to reflect the modifications from the instruction. You need to provide a final rating for the editing result from 0 to 10, with 10 being the perfect edit that precisely reflects all changes while preserving everything else. Conclude your evaluation with the score wrapped with \textless{}s\textgreater{}\dots\textless{}/s\textgreater{}. \\ \midrule

Step-1 Prompt &
Please analyze the modifications that should be reflected in the final image by strictly reasoning over the editing trace, given the source image and the complex instruction. Modifications include but are not limited to object placement, object removal, object modification, background change, and camera movement. \\ \midrule

Step-1 Input &
Source Image $I_\text{src}$, Complex Instruction $T_c$ \\ \midrule

Step-2 Prompt &
Please evaluate whether the edited image accurately reflects the modifications specified in the complex editing instruction based on your previous reasoning. Be critical of the changes made during the editing process. Conclude your assessment with a final score wrapped with \textless{}s\textgreater{}\dots\textless{}\textbackslash{}s\textgreater{}. \\ \midrule

Step-2 Input &
Generated editing result $I_\text{gen}$ \\ \midrule

Output Example &
...\textless{}s\textgreater{}5.0\textless{}/s\textgreater{} \\

\bottomrule
\end{tabular}
}
\vspace{-1em}
\end{table*}

\vspace{2mm}\noindent\textbf{Dependency Injection.} We maintain a detailed log of all executed editing operations when constructing editing chains. This log records each operation’s type, sampled parameters, and the resulting changes to object states (\eg, updated 3D coordinates). Our dependency injector builds directly on this editing history. For every newly sampled operation (excluding viewpoint and background changes), we decide whether to inject one of two inter-step dependencies, operation reference or position reference, with probability $p_d=0.5$. Concretely, the task construction begins with only operation-reference injection enabled: for every operation after the first, there is a 50\% chance that its instruction will be augmented with reference-based object descriptions. Once an object has been removed or repositioned, position-reference injection becomes available, and both dependency types are then sampled with equal probability (25\% each).

\subsection{Instruction Generation}
We use a library of hand-crafted templates to generate natural language instructions for editing operations. Each template contains placeholders that get filled with operation-specific parameters like object names, colors, directions, and action verbs. Action verbs are randomized to prevent repetitive language. For example, object addition can use "Add", "Place", or "Put" as the opening verb, object translation alternates between "Move" and "Relocate", and color changes use "Recolor" or "Paint". This randomization enables natural variations that improve instruction diversity. For inter-step dependency, (1) operation reference connects the current edit to a recently manipulated object using temporal descriptions like "the object that was moved" or "the object that was recolored", which anonymizes the object identity, using only "object" or "the object" to force models to track which object was affected by the referenced operation. When multiple objects share the same history operations (\eg, more than one objects were moved), we enrich the referenced action verb with attribute details to ensure the reference unambiguously identifies the intended target. For instance, if both a chair and a book were moved by prior edits, referring to "the object that was moved" would be ambiguous, so the instruction will be further enriched with "the object that was moved onto the desk" to distinguish between them. (2) Position reference encodes spatial memory by referring to locations where objects were originally placed. These instructions use temporal markers to clarify the reference point: "Move the lamp to where the vase originally was" or "Add a book where the cup initially was". 

\subsection{Heuristic Filters for Data Generation}
To avoid untraceable transitions that trivialize the task in single-turn settings, we heuristically filter out operation sequences whose later steps depend on transient intermediate states that can not be inferred without step-by-step visualization. For instance, a sequence that adds an object and subsequently rotates it is excluded, since the rotation relies on the object’s intermediate orientation introduced by the addition operation, which is not observable in the single-turn setting. Plus, we only allow position reference to point to the original position from scene initialization rather than intermediate positions during editing. This design ensures position references have clear, unambiguous targets.

Meanwhile, we do not explicitly exclude trajectories in which a later editing operation overwrites an earlier one, as such interactions naturally arise in complex editing tasks and contribute to instruction complexity from different angles. Moreover, correctly resolving these dependencies is itself a non-trivial challenge for achieving high-fidelity edits. We also avoid any explicit prompt rewriting that could dilute or simplify the constructed instruction complexity in our experiments. This choice allows us to directly stress-test existing models’ editing capabilities using the full, unaltered complex instructions.

\begin{table*}[t]
\centering
\caption{\textbf{Classifier-Free Guidance Implementation Comparisons}}
\label{tab:cfg_comparison}
\resizebox{\textwidth}{!}{%
\begin{tabular}{@{}l|l|l|l|l|l@{}}
\toprule
 & CFG\_IMG & CFG\_TEXT & ST\_INPUT & CTX\_INPUT & Guidance Computation \\ \midrule
INPUT: $\langle I_0, \hat{T}_c \rangle$ &
  &
  &
  &
  &
  \\ \midrule
Single-Turn Editing &
  $\hat{T}_c$ &
  $I_0$ &
  &
  &
  \begin{tabular}[c]{@{}l@{}}%
  $v_{\text{text}} = v_{\theta}(\text{cfg\_text}) + \gamma_{text}\bigl(v_{\theta}(\text{input}) - v_{\theta}(\text{cfg\_text})\bigr)$ \\
  $v = v_{\theta}(\text{cfg\_img}) + \gamma_{img}\bigl(v_{\text{text}} - v_{\theta}(\text{cfg\_img})\bigr)$
  \end{tabular} \\ \bottomrule
INPUT: $\langle I_0, T_1, I_1, \dots, T_i \rangle$ &
  &
  &
  &
  &
  \\ \midrule
Default &
  $\{T_j\}_{j=1}^i$ &
  $I_0, \{(T_j, I_j)\}_{j=1}^{i-1}$ &
  &
  &
  \begin{tabular}[c]{@{}l@{}}%
  $v_{\text{text}} = v_{\theta}(\text{cfg\_text}) + \gamma_{text}\bigl(v_{\theta}(\text{input}) - v_{\theta}(\text{cfg\_text})\bigr)$ \\
  $v = v_{\theta}(\text{cfg\_img}) + \gamma_{img}\bigl(v_{\text{text}} - v_{\theta}(\text{cfg\_img})\bigr)$
  \end{tabular} \\ \midrule
Separate &
  $\{T_j\}_{j=1}^i$ &
  $\{I_j\}_{j=0}^{i-1}$ &
  &
  &
  \begin{tabular}[c]{@{}l@{}}%
  $v_{\text{text}} = v_{\theta}(\text{cfg\_text}) + \gamma_{text}\bigl(v_{\theta}(\text{input}) - v_{\theta}(\text{cfg\_text})\bigr)$ \\
  $v = v_{\theta}(\text{cfg\_img}) + \gamma_{img}\bigl(v_{\text{text}} - v_{\theta}(\text{cfg\_img})\bigr)$
  \end{tabular} \\ \midrule
Full-Context &
  $I_0, \{(T_j, I_j)\}_{k=1}^{i-2}, \{T_k\}_{k=i-1}^i$ &
  $I_0, \{(T_j, I_j)\}_{k=1}^{i-1}$ &
  &
  &
  \begin{tabular}[c]{@{}l@{}}%
  $v_{\text{text}} = v_{\theta}(\text{cfg\_text}) + \gamma_{text}\bigl(v_{\theta}(\text{input}) - v_{\theta}(\text{cfg\_text})\bigr)$ \\
  $v = v_{\theta}(\text{cfg\_img}) + \gamma_{img}\bigl(v_{\text{text}} - v_{\theta}(\text{cfg\_img})\bigr)$
  \end{tabular} \\ \midrule
Context-Guided &
  $\{T_j\}_{j=1}^{i}$ &
  $I_0$ &
  $I_0, \{T_j\}_{j=1}^{i}$ &
  \begin{tabular}[c]{@{}l@{}}%
  $I_0, \{T_j\}_{j=1}^{i}$, \\ 
  $I_{i-1}$ or other context composition
  \end{tabular} &
  \begin{tabular}[c]{@{}l@{}}%
  $v_{\text{text}} = v_{\theta}(\text{cfg\_text}) + \gamma_{text}\bigl(v_{\theta}(\text{st\_input}) - v_{\theta}(\text{cfg\_text})\bigr)$ \\
  $v_{\text{d}} = v_{\theta}(\text{cfg\_img}) + \gamma_{img}\bigl(v_{\text{text}} - v_{\theta}(\text{cfg\_img})\bigr)$ \\
  $v = v_{\text{d}} + \gamma_{ctx}\bigl(v_{\theta}(\text{ctx\_input)} - v_{\theta}(\text{st\_input})\bigr)$
  \end{tabular} \\ \bottomrule
\end{tabular}%
}
\vspace{-1em}
\end{table*}

\subsection{Dataset Analysis}
\label{sec:dataset_analysis}
We construct a synthetic dataset including 60K independent chains and 34K dependent chains through our pipeline. Samples of constructed editing sequences are illustrated in Figure~\ref{fig:dataset_gallery_main}. In Figure~\ref{fig:dataset_analysis}, we provide a statistical analysis over the dataset from the following four aspects: (1) Editing Chain Length: The constructed editing chains have an average length of 9.98, skewed slightly towards longer sequences; (2) Dependency Ratio: Among all the editing chains with inter-step dependencies, we have dependency ratios concentrated in the range of (0, 0.55], with an average ratio of 0.29; (3) Unique Object Count: Each scene have 9.19 unique objects on average, which are drawn from 2,000+ 3D assets, offering sufficient scene complexity for editing chain construction; (4) Distribution of Editing Operations: 10 different operations are distributed in balance, mildly favoring object translation and replacement.

\section{Experimental Details}
\label{sec:exp_details}
\noindent\textbf{Baselines.} In synthetic experiments, we compare our approach against four state-of-the-art editing models as baselines: Qwen-Image~\cite{wu2025qwenimagetechnicalreport}, GPT-4o~\cite{openai2025addendum}, Gemini 2.5 Flash Image~\cite{gemini2025} and BAGEL (Zero-Shot)~\cite{deng2025emerging}. All four baselines are evaluated in a zero-shot manner. To evaluate their sequential editing performance, we implement Gemini 2.5 Flash Image and BAGEL (Zero-Shot) as in-context editing, where all subsequent edits attend to all previous contextual inputs, and GPT-4o as canonical multi-turn editing, in which every edit is performed on the result from the last editing session without access to the historical context. 

\vspace{2mm}\noindent\textbf{Implementation details.} Our framework is built on top of BAGEL~\cite{deng2025emerging}, which naively supports the training of interleaved generation. Specifically, we adapted the implementation from~\cite{Li2025ZebraCoTAD} for instruction decomposition and in-context editing training tasks. We set max\_num\_tokens to 55000 to cover more complex tasks with longer interleaved sequences, and increased the dropout rate for different tokens to \{text: 0.15, VAE: 0.4, ViT: 0.4\}. For other hyperparameters, we reuse the configurations provided by the authors and finetune the model for 12,000 steps in synthetic experiments and 1,000 steps in sim-to-real experiments, with a learning rate of 2e-5. All experiments are performed on 8 H100 GPUs. In Complex-Edit evaluation, we employ direct concatenation, rather than using GPT-4o, to compound 8 atomic instructions in each task.

\begin{table}[t]
\centering
\caption{Evaluations on different sequential editing paradigms}
\label{tab:cfg_seq_editing}
\resizebox{0.8\columnwidth}{!}{%
\begin{tabular}{@{}lll@{}}
\toprule
Sequential Editing   & Independent & Dependent \\ \midrule
Default (K=3)        & 4.06        & 4.07      \\
Default (K=5)        & \textbf{4.18}        & \textbf{4.13}      \\ \midrule
Separate (K=3)       & 4.11        & 4.12      \\
Separate (K=5)       & \textbf{4.19}        & 4.05      \\ \midrule
Full-Context (K=3)   & 4.14        & 4.06      \\
Full-Context (K=5)   & 4.16        & \textbf{4.13}      \\ \midrule
Context-Guided (K=3) & 4.15        & 4.04      \\
Context-Guided (K=5) & \textbf{4.20}        & \textbf{4.14}      \\ \bottomrule
\end{tabular}%
}
\vspace{-1em}
\end{table}

\vspace{2mm}\noindent\textbf{Metrics.} For DINOv3-based similarity metrics, we compute (1) DINO-I: the cosine similarity between the DINO features of the generated image and its reference target image as $cos(\boldsymbol{C}(I_{gen}), \boldsymbol{C}(I_{tgt}))$; (2) DINO-D: the cosine similarity over editing directions as $cos(\boldsymbol{C}(I_{gen}) - \boldsymbol{C}(I_{src}), \boldsymbol{C}(I_{tgt}) - \boldsymbol{C}(I_{src}))$, where $\boldsymbol{C}$ denotes DINOv3 encoder, and $I_{src}, I_{tgt}, I_{gen}$ denote the source image, the target image and the generated image respectively. For GPT-5–based evaluation, we adopt a two-step procedure: (a) GPT-5 first infers the desired modifications by jointly analyzing the source image and the complex instruction, (b) we then provide GPT-5 with the generated editing result to assess how well it satisfies those inferred requirements. The prompts we applied are shown in Table~\ref{tab:eval_prompts}.

\section{More Experiments and Ablation Studies}
\noindent\textbf{Classifier-Free Guidance.} In our framework, different ways of computing classifier-free guidance encompass versatile editing paradigms. We instantiate our framework with BAGEL~\cite{deng2025emerging}, in which the authors provide one way to compute classifier-free guidance that is slightly different from what we used in full-context sequential editing and context-guided sequential editing. In the following, we term the editing paradigm induced by the default CFG computation from BAGEL as ``default sequential editing''. Specifically, the sequential editing performances of BAGEL (Zero-Shot) in Table~\ref{tab:independent_chain_eval} and Table~\ref{tab:dependent_chain_eval}, BAGEL-Zero-Shot (2-Step Sequential Editing), as well as BAGEL-R (2-Step Sequential Editing) in Table~\ref{tab:sim_to_real}, are evaluated with default sequential editing paradigm. We provide a detailed comparison between default, full-context, context-guided paradigms, along with another editing paradigm which separately compute the guidance from text-only and image-only sequences (termed "separate") in Table~\ref{tab:cfg_comparison}, following the notations from BAGEL's implementation. The default, separate and full-context sequential editing paradigms utilize all contextual inputs to generate the next editing result with different context focuses, whereas context-guided editing selectively use previous editing results as an additional signal to guide single-turn editing. Furthermore, we report the evaluation results on both independent and dependent chains for all four sequential editing paradigms with our finetuned BAGEL in Table~\ref{tab:cfg_seq_editing}. We discover that all sequential editing paradigms achieves better performance than single-turn editing after finetuning, highlighting the effectiveness of sequential decomposition and robustness of our framework. Also, the default and full-context editing paradigms generally have similar performances on the challenging dependent setting, where the separate editing paradigm encounters a decline when the number of decomposition steps increases. Context-guided editing overall can be considered as a robust paradigm for sequential decomposition.

\vspace{2mm}\noindent\textbf{Effect of $\gamma_{ctx}$.} By default, we use $\gamma_{ctx} = 2.5$ for synthetic editing tasks and $\gamma_{ctx} = 0.5$ for real-world editing tasks. We also test context-guided sequential editing ($K=3$) with different values of $\gamma_{ctx}$ on 100 unseen independent and dependent chains, and provide the results in Table~\ref{tab:ablation_gamma_ctx}. We found $\gamma_{ctx} = 2.5$ achieves the best overall performance among the evaluated settings.

\begin{table}[h]
\centering
\vspace{-0.5em}
\caption{\textbf{Ablation on $\gamma_{ctx}$ for Context-Guided Editing (K=3)}}
\label{tab:ablation_gamma_ctx}
\resizebox{0.9\columnwidth}{!}{%
\begin{tabular}{@{}l|llll@{}}
\toprule
$\gamma_{ctx}\ (K=3)$    & 0.5  & 1.5  & 2.5  & 3.5  \\ \midrule
Independent Chains (100) & \textbf{4.41} & 4.24 & \textbf{4.35} & 4.30  \\
Dependent Chains (100)   & 3.96 & 4.10  & \textbf{4.23} & 3.87 \\ \bottomrule
\end{tabular}%
}
\vspace{-1em}
\end{table}

\vspace{2mm}\noindent\textbf{Effect of context composition.} By default, we use the last editing result alone as the guidance for guidance computation in context-guided editing. We also compare different context compositions on 100 independent and dependent chains, and provide the results in Table~\ref{tab:ablation_ctx_composition}. We observe that using all previous editing results can lead to a performance drop. We hypothesize that the errors in all previous editing results can cumulatively impact the current edits, leading to compounding errors and performance degradation.
\begin{table}[h]
\vspace{-0.5em}
\caption{\textbf{Ablation on context composition for Step $i$}}
\label{tab:ablation_ctx_composition}
\resizebox{\columnwidth}{!}{%
\begin{tabular}{@{}lll@{}}
\toprule
\begin{tabular}[c]{@{}l@{}}Context Composition \\ (K=3, $\gamma_{ctx} = 2.5$)\end{tabular} &
  \begin{tabular}[c]{@{}l@{}}All Previous Editing Results \\ $\{I_{j}\}_{j=0}^i$\end{tabular} &
  \begin{tabular}[c]{@{}l@{}}Last Editing Result \\ $I_{j-1}$\end{tabular} \\ \midrule
Independent Chains (100) & 4.18 & \textbf{4.35} \\
Dependent Chains (100) & 4.13 & \textbf{4.23} \\ \bottomrule
\end{tabular}%
}
\vspace{-1em}
\end{table}

\vspace{2mm}\noindent\textbf{Ablation on decomposition steps.} We study how the number of decomposition steps impacts the performance of context-guided sequential editing and provide the results in Table~\ref{tab:ablation_decomposition_step}. We observe that more decomposition steps yield performance gains. Specifically, $K=4$ performs the best on independent chains, while $K=5$ achieves the highest performance on editing tasks with inter-step dependencies.
\begin{table}[h]
\vspace{-0.5em}
\caption{\textbf{Ablation on decomposition steps (K) for Context-Guided Sequential Editing with $\gamma_{ctx} = 2.5$}}
\label{tab:ablation_decomposition_step}
\resizebox{\columnwidth}{!}{%
\begin{tabular}{@{}llllll@{}}
\toprule
$K$                & Single-Turn (1) & 2    & 3    & 4    & 5    \\ \midrule
Independent Chains & 4.0             & 4.06 & 4.15 & \textbf{4.27} & 4.20 \\
Dependent Chains   & 3.91            & 4.11 & 4.04 & 4.11 & \textbf{4.14} \\ \bottomrule
\end{tabular}%
}
\vspace{-1em}
\end{table}

\vspace{2mm}\noindent\textbf{Editing sequences versus pairwise transitions as training data.} We also investigate how the training data formulation affects the sequential editing performance. To enable sequential editing, we can either train our model on editing sequences with objectives described in~\ref{sec:method}, or break each editing sequence down into individual editing pairs and optimize the model to perform single-turn editing on these pairwise transition data. We train BAGEL on these two types of data with their corresponding objectives respectively, and compare their sequential editing performance on independent chains with three decomposition steps. Table~\ref{tab:sequential_vs_pairwise} shows that even though the model is trained on data that teaches the model how to perform the visual transitions individually, it remains difficult to perform sequential editing robustly due to the train-test gap. On the other hand, training on editing sequences naturally encourages the model to perform sequential editing by utilizing the rich contextual inputs, leading to better performance.

\begin{table}[h]
\centering
\vspace{-0.5em}
\caption{\textbf{Comparison between sequential and pairwise training data for sequential editing inference}}
\label{tab:sequential_vs_pairwise}
\resizebox{0.7\columnwidth}{!}{%
\begin{tabular}{@{}l|ll@{}}
\toprule
\# Editing Operations in Tasks & 3   & \textgreater{}3 \\ \midrule
Pairwise Transitions             & 2.94 & 2.30            \\
Editing Sequences          & 5.03 & 4.04            \\ \bottomrule
\end{tabular}%
}
\vspace{-1em}
\end{table}

\vspace{2mm}\noindent\textbf{Ablation on single-turn editing training objective.} We investigate whether incorporating a single-turn editing objective during training affects the performance of context-guided editing. In Table~\ref{tab:ablation_single_turn_edting_objective}, we compare context-guided editing results using three decomposition steps (\ie $K=3$) on independent chains, with and without the single-turn editing objective enabled. We find that the instruction decomposition and in-context editing objectives alone already yield robust improvements through context-guided sequential editing, while adding the single-turn editing objective provides a further performance boost.

\begin{table}[h]
\centering
\vspace{-0.5em}
\caption{\textbf{Impact of single-turn editing training objective on context-guided sequential editing}}
\label{tab:ablation_single_turn_edting_objective}
\resizebox{\columnwidth}{!}{%
\begin{tabular}{@{}lll@{}}
\toprule
Training Objective    & w/o Single-Turn Editing & w/ Single-Turn Editing \\ \midrule
Context-Guided Editing & 4.12                    & 4.15                   \\ \bottomrule
\end{tabular}%
}
\vspace{-1em}
\end{table}

\vspace{2mm}\noindent\textbf{Additional Baselines on Complex-Edit.} In Table~\ref{tab:additional_real_baselines}, we re-run AnyEdit~\cite{yu2024anyedit} and UltraEdit~\cite{Zhao2024UltraEditIF} with instructions composed by concatenating 8 atomic editing operations.

\begin{table}[h]
\centering
\vspace{-0.5em}
\caption{\textbf{Additional baselines on Complex-Edit}}
\label{tab:additional_real_baselines}
\resizebox{0.95\columnwidth}{!}{%
\begin{tabular}{@{}llll@{}}
\toprule
Complex-Edit    & IF $\uparrow$ & IP $\uparrow$ & PQ $\uparrow$\\ \midrule
AnyEdit (Single-Turn) & 1.68   & 8.49 & 7.37                   \\ 
AnyEdit (2-Step Sequential Editing) & 2.61   & 7.43 & 6.63                   \\ 
UltraEdit (Single-Turn) & 6.00   & 5.49 & 7.19                   \\ 
UltraEdit (2-Step Sequential Editing) & 5.78   & 5.01 & 7.38     \\ \bottomrule
\end{tabular}%
}
\vspace{-1em}
\end{table}

\vspace{2mm}\noindent\textbf{Inference Cost.} In Table~\ref{tab:inference_cost}, we analyze inference cost on a single A100 GPU with \textit{K}=$\{$1,3,5$\}$ and 50 denoising steps per image. We find that fully generating \textit{K}\textgreater 1 images with CGSE takes approximately 0.85\textit{K} times the cost of single-turn editing (\textit{K}=1)\textemdash a 15\% speedup enabled by KV cache. However, this cost can be further reduced on more advanced GPUs (\eg H200) and with faster sampling after distillation.

\begin{table}[h]
\centering
\vspace{-0.5em}
\caption{\textbf{Inference cost analysis}}
\label{tab:inference_cost}
\resizebox{\linewidth}{!}{%
\begin{tabular}{@{}lccc@{}}
\toprule
Decomposed Steps (\textit{K})          & 1    & 3            & 5            \\ \midrule
Inference Time Per Task (s) & 43.2 $\pm$ 1.5 & 112.1 $\pm$ 1.6 & 182.0 $\pm$ 1.4 \\ \bottomrule
\end{tabular}%
}
\vspace{-1.3em}
\end{table}

\section{More Related Work}

\noindent\textbf{Instruction-based Image Editing.} As opposed to prior works~\cite{hertz2022prompt, couairon2022diffedit, kawar2023imagic, meng2021sdedit, cao2023masactrl, tumanyan2023plug} that rely on detailed textual descriptions of the desired image to guide the editing process, ``instructions'' specify how the image should be changed, providing a user-friendly interface for image editing. Fueled by advancements in model architectures and multimodal datasets for image synthesis, instruction-based image editing has achieved rapid progress over recent years. InstructPix2Pix~\cite{Brooks2022InstructPix2PixLT} pioneered the use of instructions for image editing by finetuning Stable Diffusion~\cite{Rombach2021HighResolutionIS} on synthetic data~\cite{hertz2022prompt}, and inspired a line of following works in which editing models are further improved to perform diversified types of edits on both real and synthetic images~\cite{Zhao2024UltraEditIF, Zhang2023MagicBrushAM, cao2025instruction, huang2024smartedit, fu2023guiding, sheynin2024emu}.

\section{More Qualitative Results}
\label{sec:more_qualitative}
\begin{figure*}[t]
    \centering
    \includegraphics[width=\linewidth]{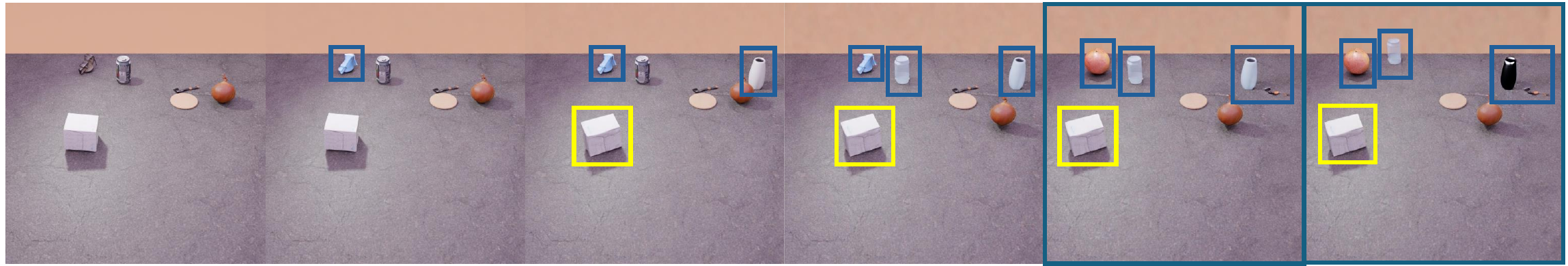}
    \caption{Full illustration of 5-step decomposition for Context-Guided Editing. %
    Decomposed Instruction over 13 editing operations: (1) Fully cover the bench vice with metal material, fully cover the bench vice with plastic material. (2) Add a vase on the floor near the yellow onion, rotate the tissue box 180 degrees, {\color{Dandelion}rotate the tissue box 90 degrees counter-clockwise}. (3) Move the yellow onion forward, fully cover the can with glass material, fully cover the vase with glass material. (4) Replace the bench vice with a grapefruit, move the hatchet to the right, pan the camera right. (5) Fully cover the vase with metal material, move the can backward. Numbered list denotes the decomposition steps. Blue boxes indicate the correct edits, and yellow boxes indicate edits that are not fully executed.}
    \label{fig:2117_5_step_full}
    \vspace{-1em}
\end{figure*}

\begin{figure*}[t]
    \centering
    \includegraphics[width=\linewidth]{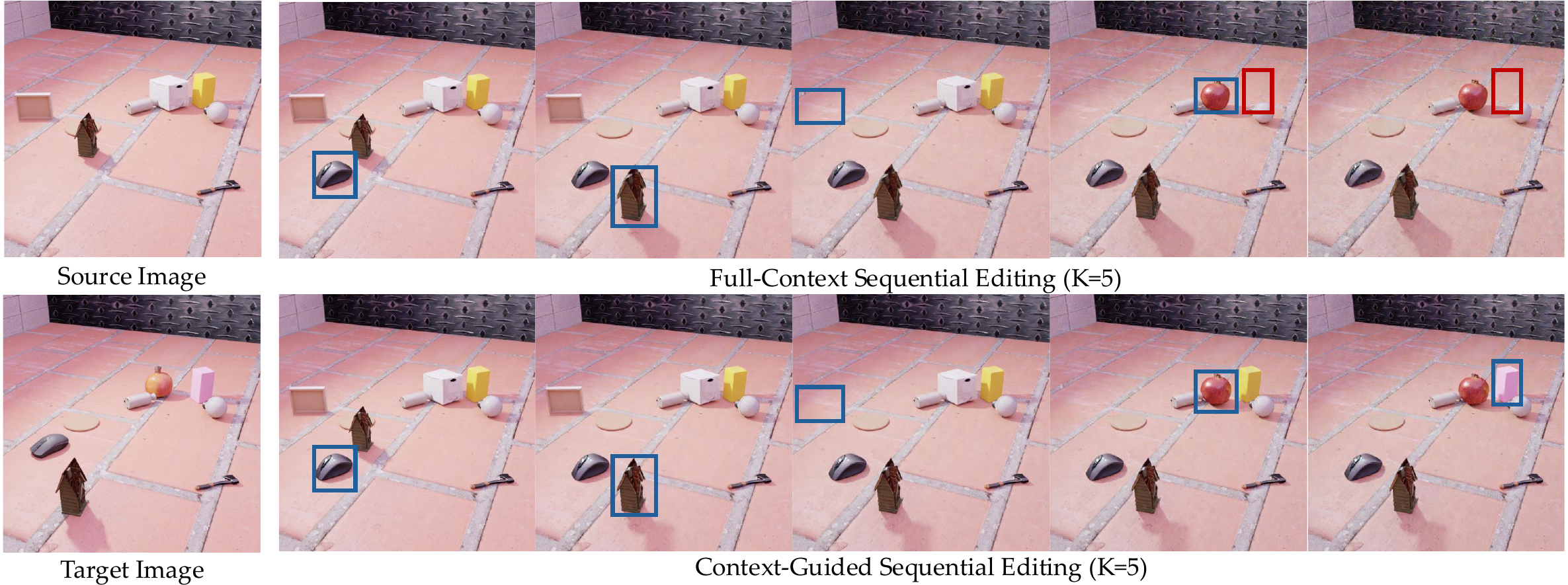}
    \caption{Full illustration of 5-step decomposition for tasks with 5 editing operations. Instruction: (1) Add a mouse on the floor near the trivet, (2) move the bird house forward, (3) remove the frame, (4) replace the tissue box with a pomegranate, and (5) paint the cuboid wood block pink. Blue boxes indicate correct edits, whereas red boxes indicate undesired ones. Numbered list denotes the decomposition steps.}
    \label{fig:2284_5_step_full}
\end{figure*}

\begin{figure*}[t]
    \centering
    \includegraphics[width=\linewidth]{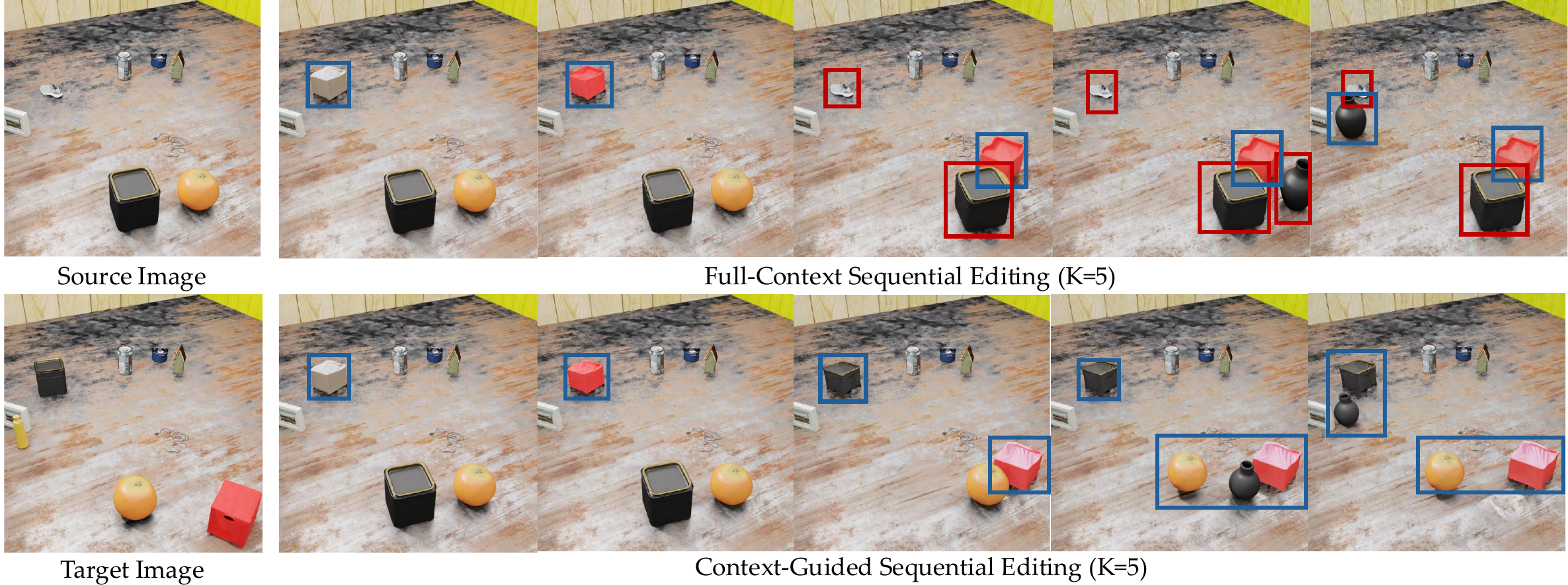}
    \caption{Full illustration of 5-step decomposition for tasks with 7 editing operations and inter-step dependencies. Instruction: (1) Replace the shoe with a tissue box, (2) paint the object that was recently placed in exchange for the shoe to red, (3) move the object that was recolored near the tangerine, move the coffee canister where the shoe initially was, (4) relocate the tangerine back to where the coffee canister initially was, add a vase where the tangerine initially was, and (5) relocate the object that was added near the frame. Blue boxes indicate correct edits, whereas red boxes indicate undesired ones. Numbered list denotes the decomposition steps.}
    \label{fig:3081_5_step_full}
\end{figure*}

\begin{figure*}[t]
    \centering
    \includegraphics[width=\linewidth]{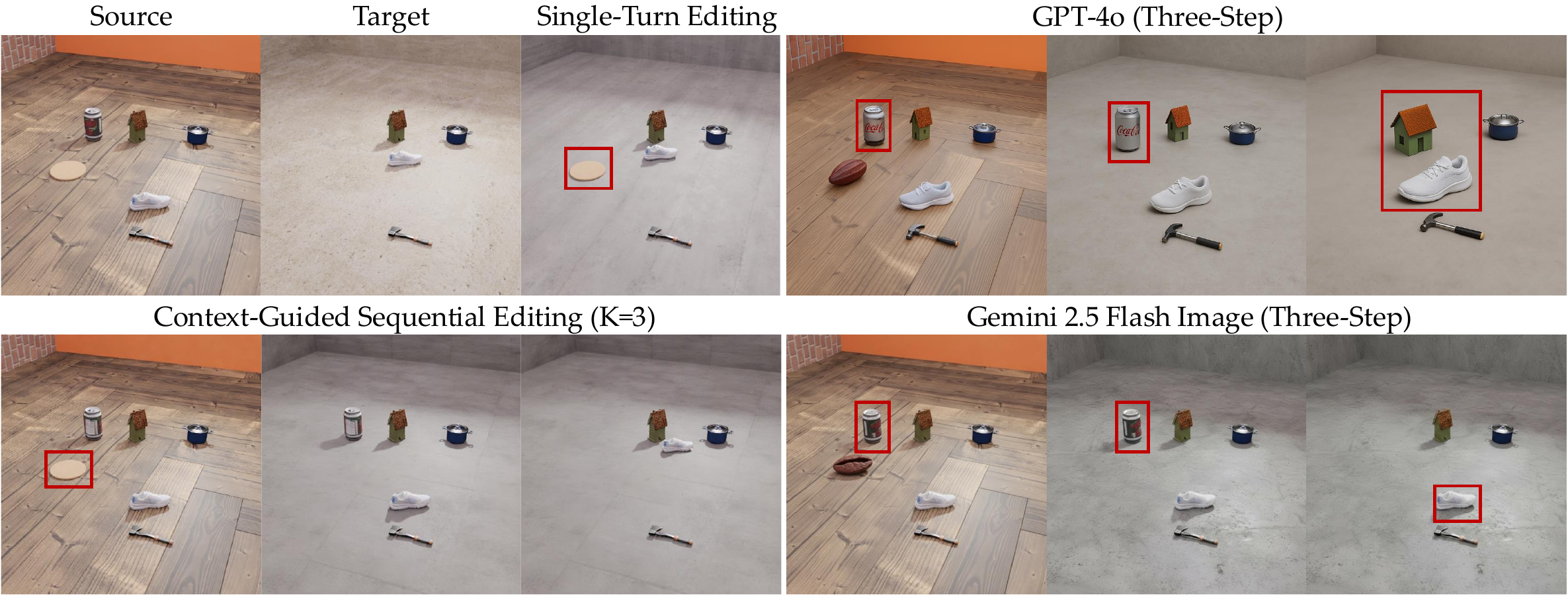}
    \caption{Comparison for tasks with 6 editing operations. Instruction: (1) Replace the trivet with a cocoa bean, rotate the can 90 degrees counter-clockwise. (2) Remove the cocoa bean, change the floor and walls to concrete. (3) Remove the can, move the shoe near the bird house. ``Single-Turn Editing'' refers to the single-turn editing result from finetuned BAGEL. Red boxes indicate wrong edits. Numbered list denotes the decomposition steps.}
    \label{fig:423_comparison}
\end{figure*}

\begin{figure*}[t]
    \centering
    \includegraphics[width=\linewidth]{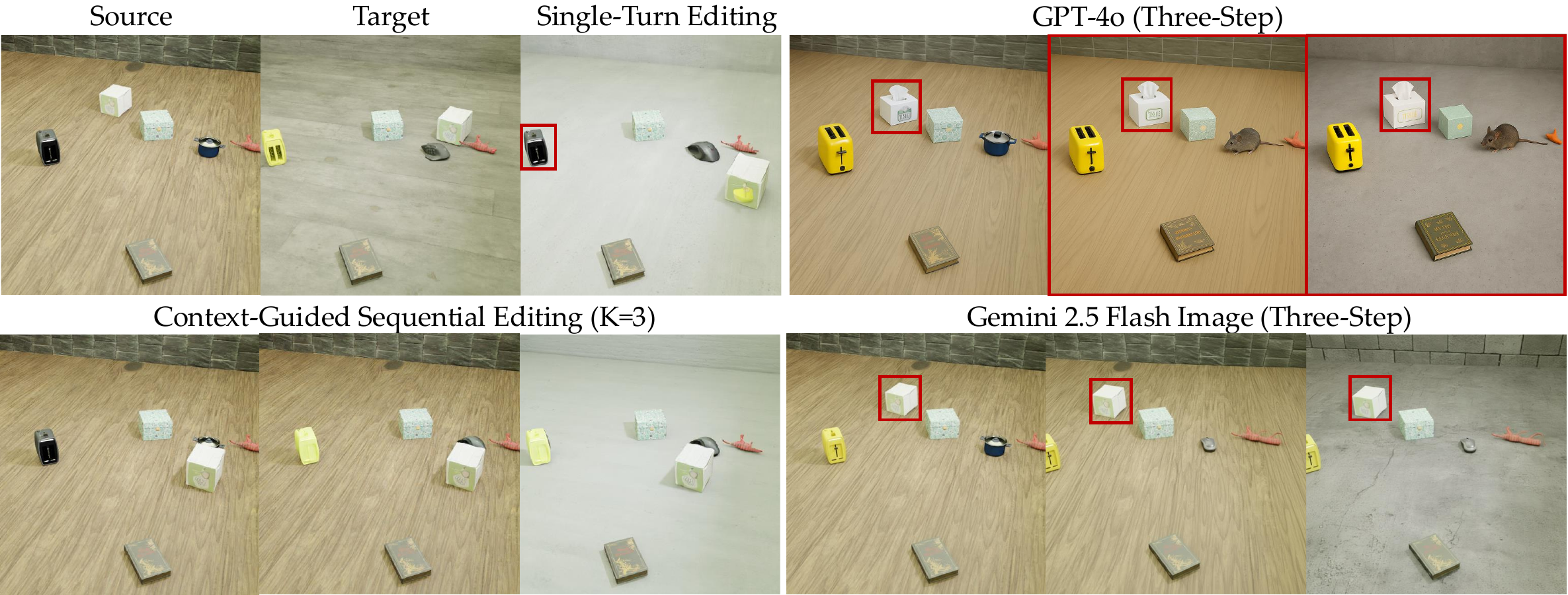}
    \caption{Comparison for tasks with 5 editing operations. Instruction: Move the tissue box near the pot, paint the toaster yellow, replace the pot with a mouse, pan the camera right, and change the floor and walls to concrete. ``Single-Turn Editing'' refers to the single-turn editing result from finetuned BAGEL. Red boxes indicate wrong edits.}
    \label{fig:1238_comparison}
\end{figure*}

\begin{figure*}[t]
    \centering
    \includegraphics[width=\linewidth]{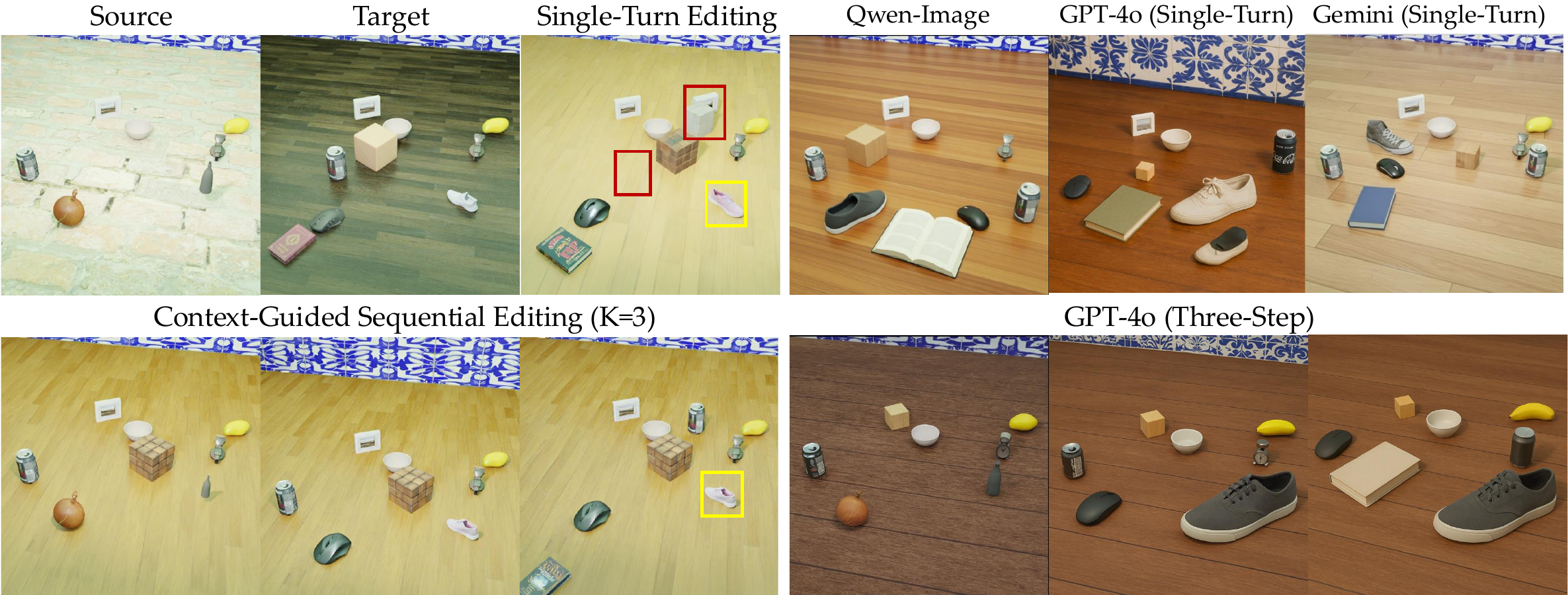}
    \caption{Comparison for tasks with 10 editing operations and inter-step dependencies. Instruction: Make the vase smaller, add a cube wood block on the floor near the mixing bowl, change the floor to wood floor, replace the vase with a shoe, tilt the camera up, {\color{Dandelion}change the material of the object that was placed in exchange for the vase to fabric material}, replace the yellow onion with a mouse, add a book on the floor near the mouse, tilt the camera down, and move the can to the right. ``Single-Turn Editing'' refers to the single-turn editing result from finetuned BAGEL. On the left, red boxes indicate wrong edits, whereas yellow boxes indicate edits that are not fully executed. Results from the proprietary baselines generally all fail to edit the image accurately.}
    \label{fig:2279_comparison}
\end{figure*}

\begin{figure*}[t]
    \centering
    \includegraphics[width=\linewidth]{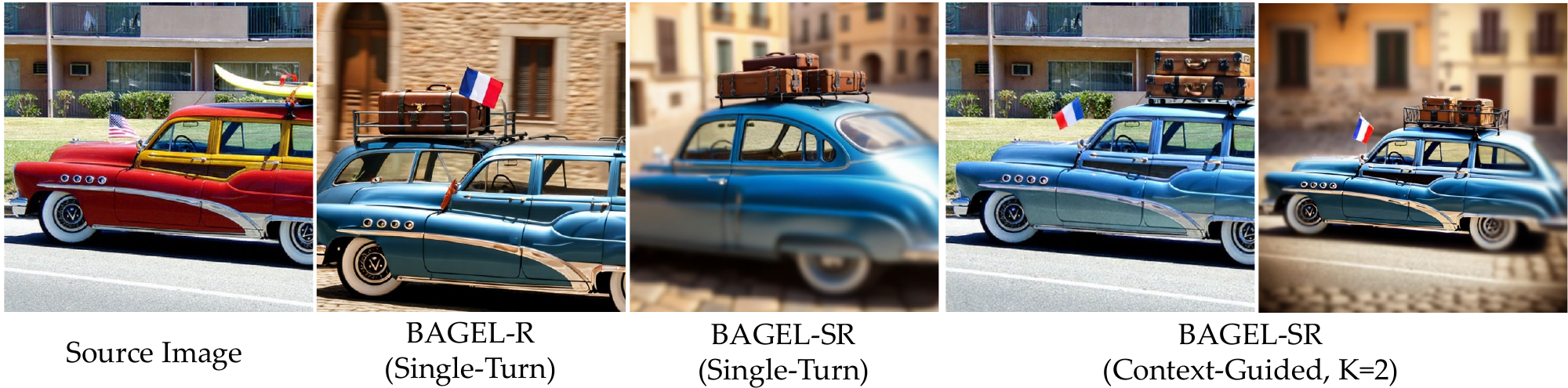}
    \caption{Comparison for real-world editing with instruction: Change the color of the car from red to metallic blue, remove the surfboard on the car, add a luggage rack with vintage suitcases on top of the car, change the small american flag to a small french flag mounted on the same spot, replace the apartment building background with a european cobblestone street, apply a warmer lighting effect, add a soft motion blur to the surrounding environment, and crop the image to emphasize the car.}
    \label{fig:81_real_comparison}
\end{figure*}

\begin{figure*}[t]
    \centering
    \includegraphics[width=\linewidth]{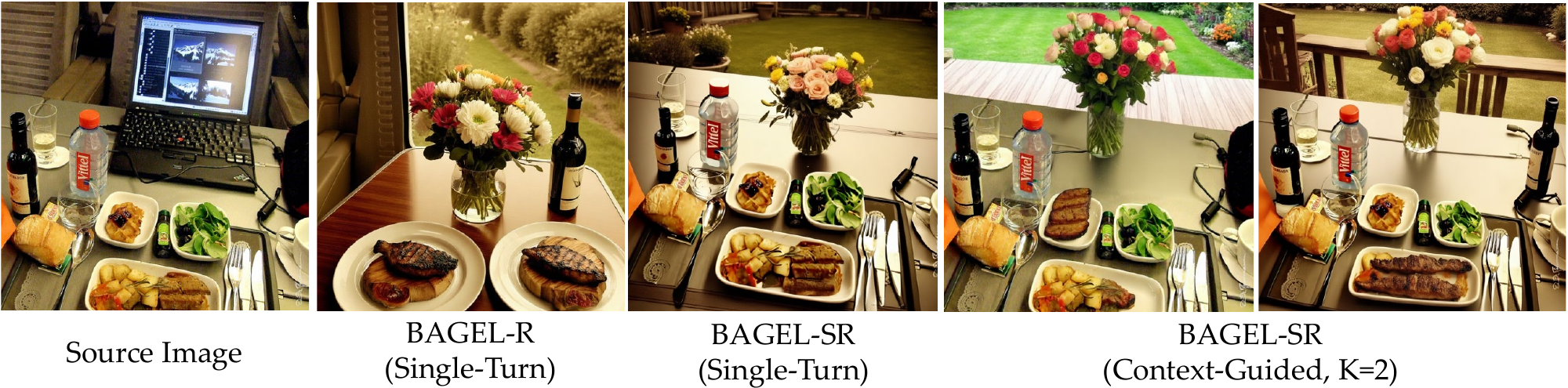}
    \caption{Comparison for real-world editing with instruction: Remove the laptop from the setup, insert a bouquet of fresh flowers in place of the laptop, replace the train seat with a wooden table and an outdoor garden background, relocate the plate with grilled meat to the center, duplicate the bottle of wine and place the copy on the opposite side of the table, add warm sunlight to simulate a late afternoon ambiance, refocus composition on the plates and enhance surrounding elements, and apply a soft sepia filter.}
    \label{fig:85_real_comparison}
\end{figure*}

\begin{figure*}[t]
    \centering
    \includegraphics[width=\linewidth]{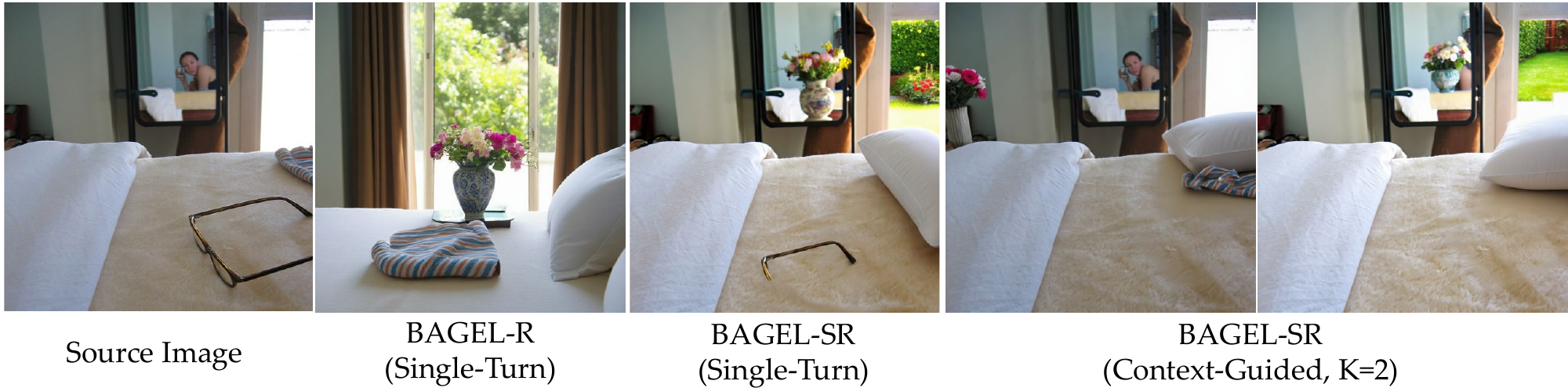}
    \caption{Comparison for real-world editing with instruction: Place a decorative vase with flowers on the table in the background, remove the glasses from the bed, reposition the striped fabric so it is neatly folded next to the location of the removed glasses, insert an extra pillow onto the bed right side, brighten the lighting slightly, add a soft glow effect, replace the visible outside portion in the mirror to show a garden view instead, and center the image frame on the bed and balance added elements.}
    \label{fig:92_real_comparison}
\end{figure*}

\begin{figure*}[t]
    \centering
    \includegraphics[width=\linewidth]{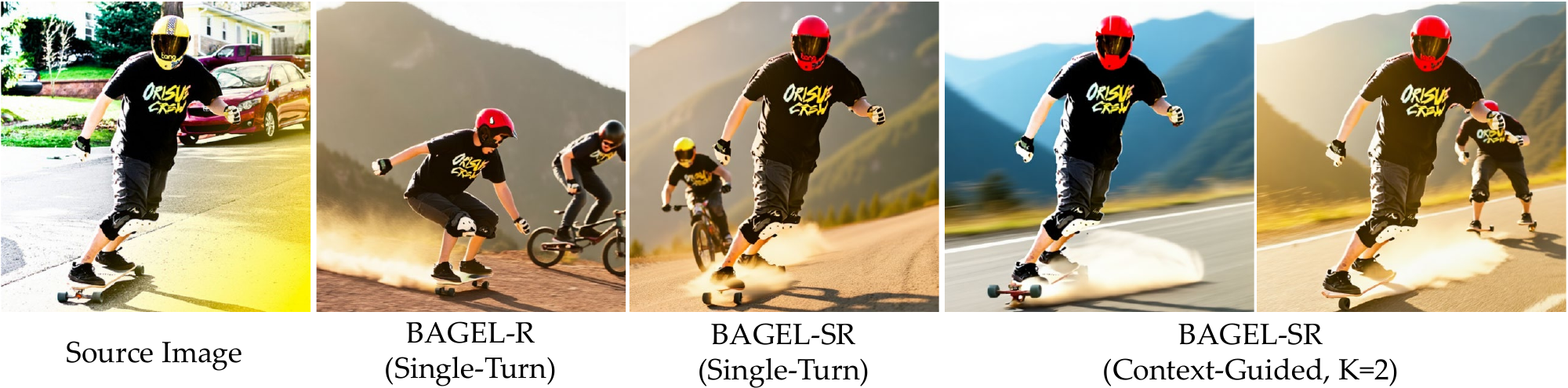}
    \caption{Comparison for real-world editing with instruction: Add a motion blur effect to the skateboard, replace the background with a mountainous environment, change the rider's helmet color to bright red, add a trail of dust particles behind the skateboard, add another rider, similarly equipped, in the image, slightly behind the main subject, adjust the duplicated rider’s pose to a crouching position, apply soft lighting from the upper-left corner, and apply a warm golden evening filter.}
    \label{fig:14_real_comparison}
\end{figure*}

\begin{figure*}[t]
    \centering
    \includegraphics[width=\linewidth]{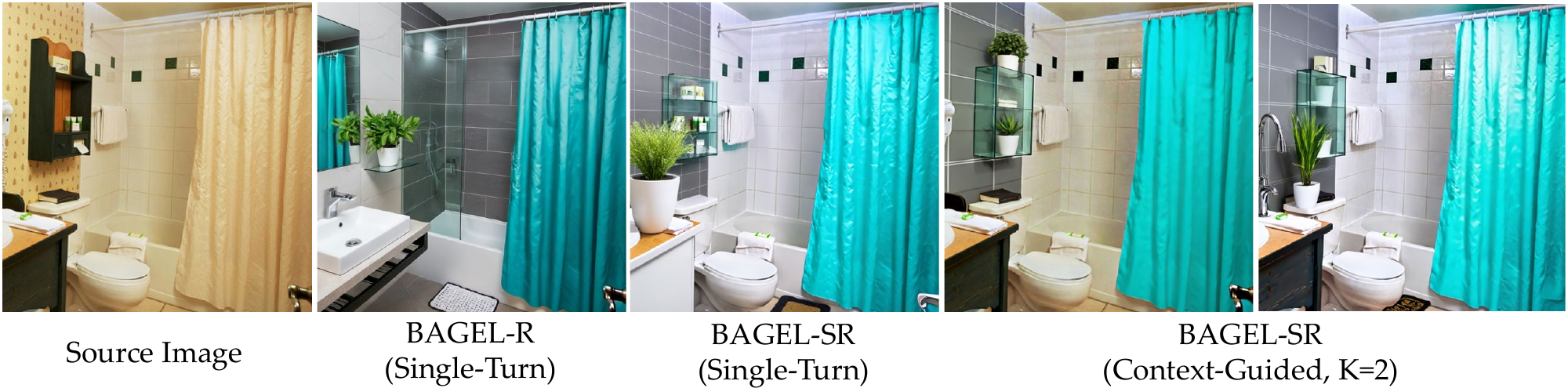}
    \caption{Comparison for real-world editing with instruction: Replace the wallpaper with a modern solid gray tile pattern, change the current wooden cabinet to a sleek, glass-door shelf, change the shower curtain to a turquoise color, add a decorative plant in a white pot on the countertop, replace the bathroom sink tap with a modern chrome tap, increase the lighting brightness, place a small anti-slip rug on the floor in front of the sink, and apply a photo filter to enhance details and colors.}
    \label{fig:181_real_comparison}
\end{figure*}

\begin{figure*}[t]
    \centering
    \includegraphics[width=\linewidth]{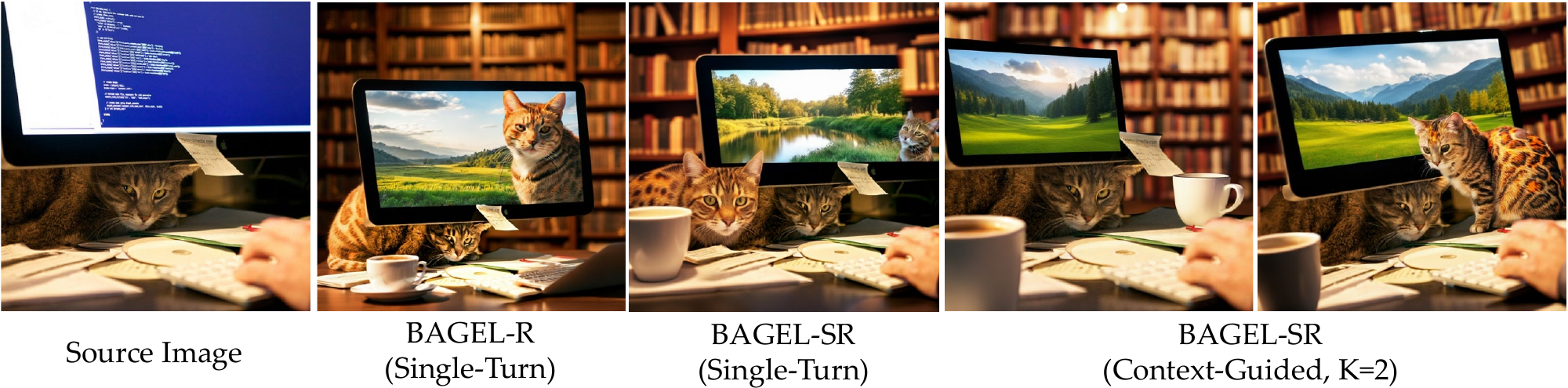}
    \caption{Comparison for real-world editing with instruction: Replace the background with an aesthetic library study environment, change the screen content to a natural scenery image, place a coffee cup near the keyboard, add a warm ambient light effect, reposition the cat to be seated beside the display, duplicate the cat and reposition it in the composition, modify the cats' fur to include patterned orange highlights, and crop the image to focus on the computer setup and the cat.}
    \label{fig:135_real_comparison}
\end{figure*}

\begin{figure*}[t]
    \centering
    \includegraphics[width=\linewidth]{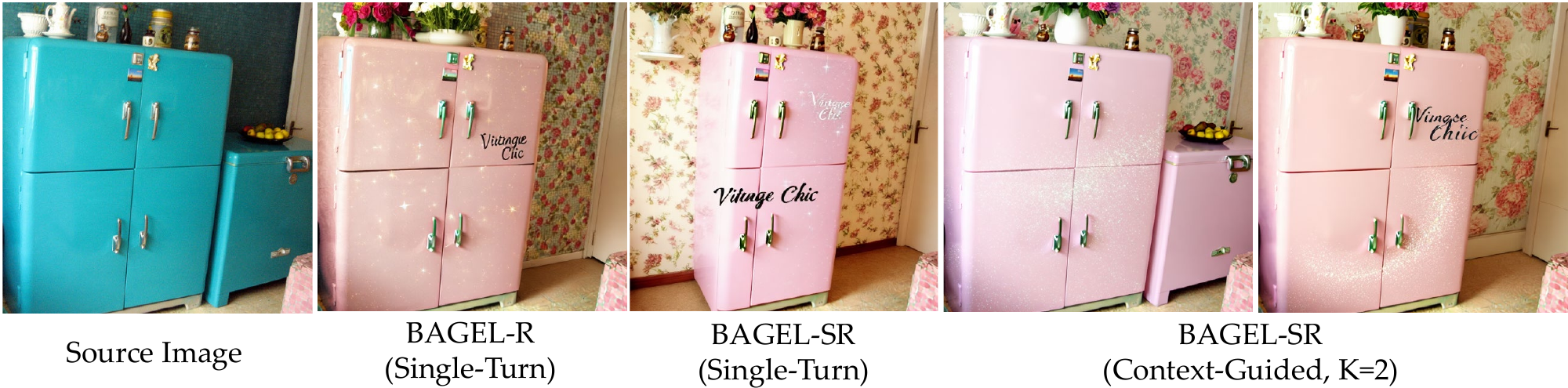}
    \caption{Comparison for real-world editing with instruction: Change the color of the refrigerator to pastel pink, add sparkles to the refrigerator's surface, place a vase with fresh flowers on top of the refrigerator, modify the wall behind the refrigerator to a floral-patterned wallpaper, remove the appliance to the right of the refrigerator, make the refrigerator appear taller by about 20\% in height, apply a warm color tone filter, and add the word 'vintage chic' in a cursive font on the refrigerator door.}
    \label{fig:201_real_comparison}
\end{figure*}

\begin{figure*}[t]
    \centering
    \includegraphics[width=\linewidth]{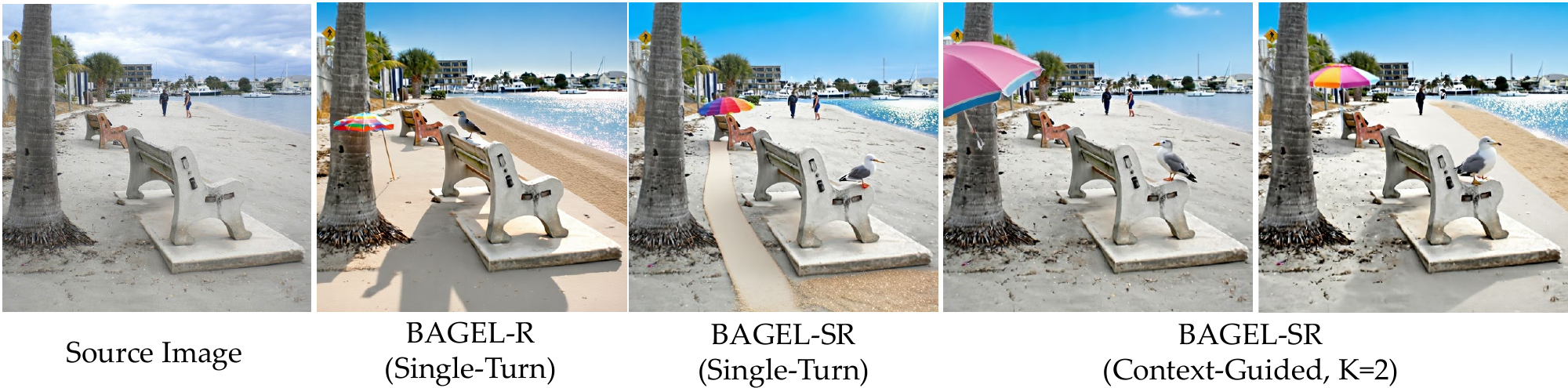}
    \caption{Comparison for real-world editing with instruction: Insert a colorful beach umbrella to the left side of the image, place a seagull perched on the foremost bench, enhance lighting to resemble a sunny day, replace the overcast background with a clear sky, add subtle sparkling effects to the water, add a sandy pathway leading to the benches, scatter light sand particles near the foreground of the image, and adjust the angle so the benches align symmetrically in the middle third of the view.}
    \label{fig:180_real_comparison}
\end{figure*}

\begin{figure*}[t]
    \centering
    \includegraphics[width=\linewidth]{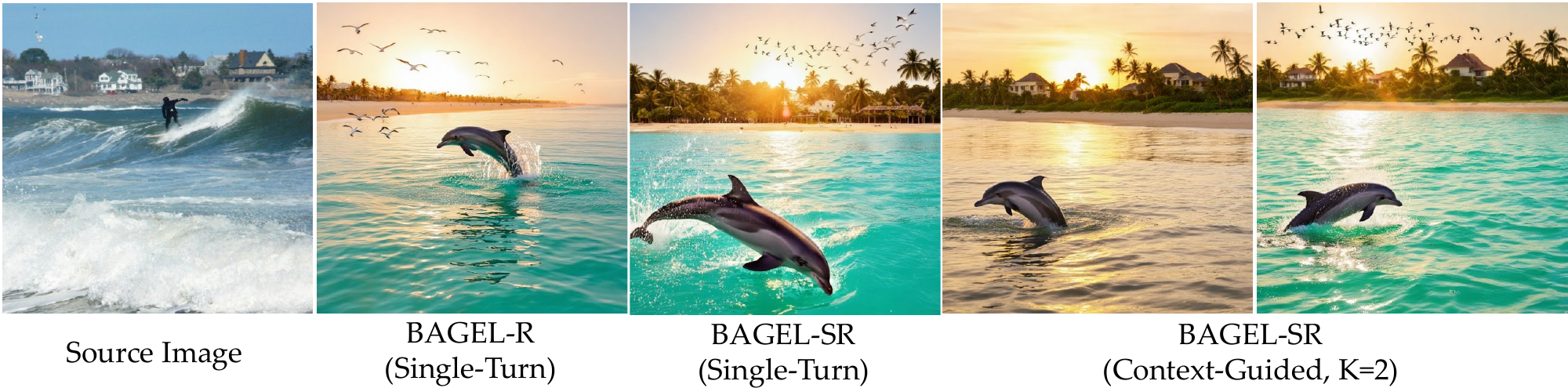}
    \caption{Comparison for real-world editing with instruction: Replace the surfer with a dolphin breaching out of the water, change the background houses and coastline to a tropical beach setting, add a setting sun low on the horizon in the background, apply a golden hour lighting filter over the image, add a flock of seagulls to the sky at a distance, change the water color to sparkling turquoise hues, add water sprays around the dolphin, and adjust shadows and highlights.}
    \label{fig:493_real_comparison}
\end{figure*}

\begin{figure*}[t]
    \centering
    \includegraphics[width=\linewidth]{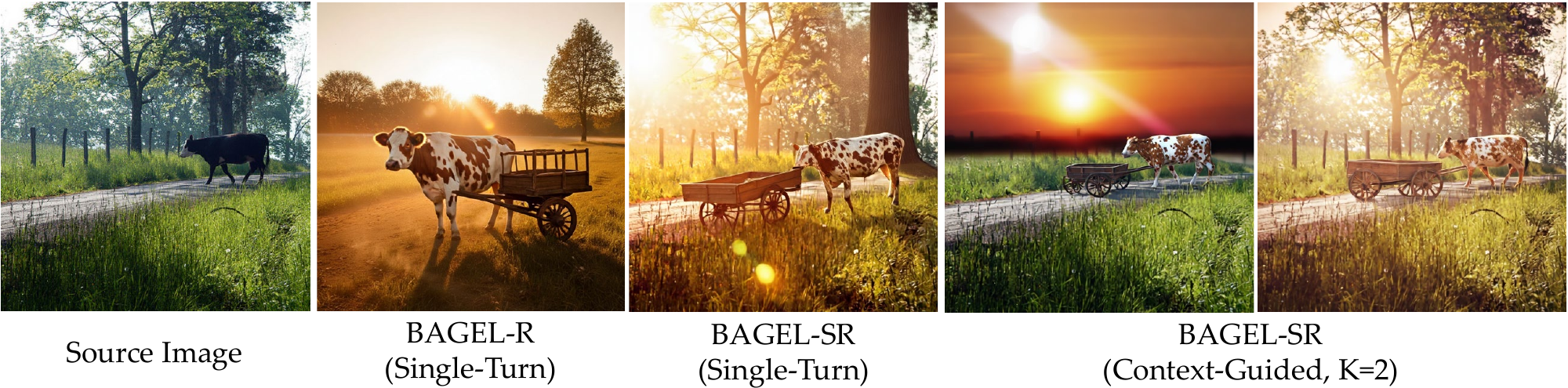}
    \caption{Comparison for real-world editing with instruction: Replace the background with a sunset scene, add a lens flare effect near the sunset, change the color of the cow to spotted brown and white, add a vintage wooden cart being pulled by the cow, add soft shadows and warm lighting around the cow and cart, add small dust particles to the background, make the tree on the right taller, and apply a golden-hue filter.}
    \label{fig:158_real_comparison}
\end{figure*}

\begin{figure*}[t]
    \centering
    \includegraphics[width=\linewidth]{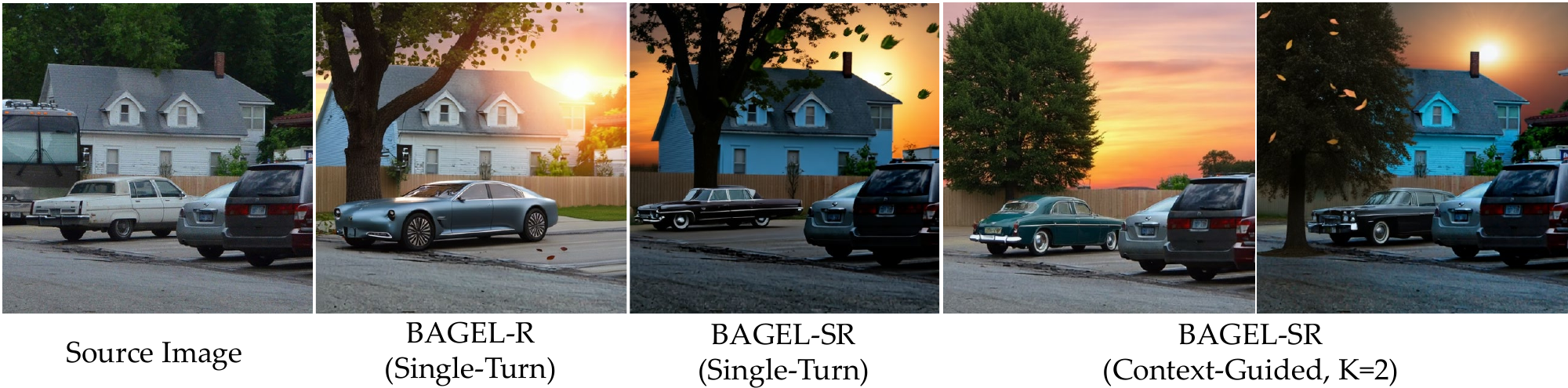}
    \caption{Comparison for real-world editing with instruction: Insert a large tree to the front left of the image, remove the bus on the left side of the image, replace the white car in the foreground with a vintage electric sedan, switch the background to depict a sunset scene, modify the color of the house to light blue, introduce floating leaves around the tree, dim the overall lighting, and add a light glow effect near the sun.}
    \label{fig:301_real_comparison}
\end{figure*}

\begin{figure*}[t]
    \centering
    \includegraphics[width=\linewidth]{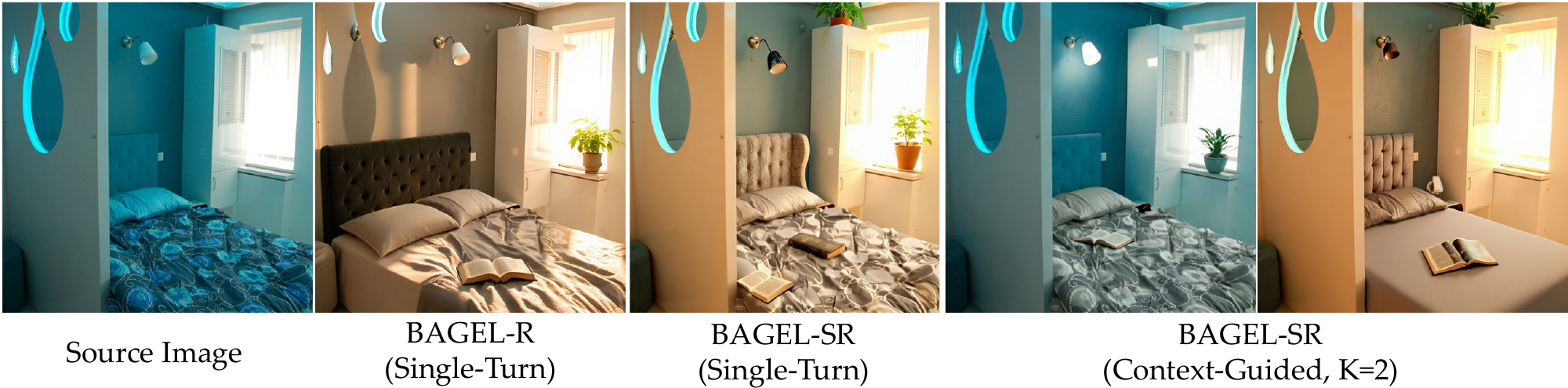}
    \caption{Comparison for real-world editing with instruction: Change the bed linens color from blue to neutral gray, place an open book on the side of the bed, add a houseplant in a ceramic pot on top of the cabinet, increase the brightness of the light from the wall sconce, alter the bed's headboard fabric to textured velvet, shift the side lamp to the left of the bed, apply a warmer tone filter to the image, and introduce sunlight rays coming through the window.}
    \label{fig:161_real_comparison}
\end{figure*}

\begin{figure*}[t]
    \centering
    \includegraphics[width=\linewidth]{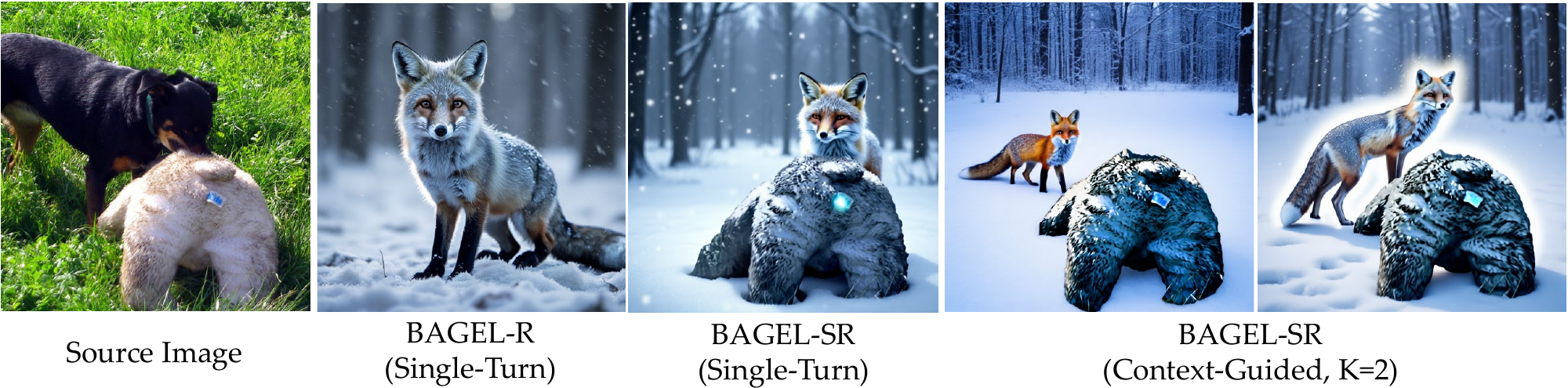}
    \caption{Comparison for real-world editing with instruction: Replace the dog with a fox, replace the grass background with a snowy forest scene, change the sheep's texture to resemble a rocky shape, modify the lighting to simulate a cold, wintry evening, introduce a snowfall effect, add a faint magical glow around the fox, alter the fox's color to a striking silver shade, and reframe the composition to focus on the fox and the modified object.}
    \label{fig:49_real_comparison}
\end{figure*}

\begin{itemize}
    \item We provide a full five-step illustration of the context-guided sequential editing on a dependent chain with 13 editing operations %
    in Figure~\ref{fig:2117_5_step_full}. Context-guided sequential editing modifies the scene following the per-step instruction, leading to high-accuracy edits with merely one incomplete modification (\ie, rotated the tissue box for 180\textdegree{} but did not continue the rotation for 90\textdegree{} counter-clockwise), which is labeled by the yellow box.
    \item We compare context-guided sequential editing with full-context sequential editing on an independent chain (Figure~\ref{fig:2284_5_step_full}) and a dependent chain (Figure~\ref{fig:3081_5_step_full}) when decomposing the complex instruction into 5 steps. We observe that context-guided sequential editing robustly produces correct edits, whereas full-context sequential editing often leads to undesired modifications.
    \item In Figure~\ref{fig:423_comparison} and Figure~\ref{fig:1238_comparison}, we compare our methods against baselines on two independent chains with 6 and 5 editing operations, respectively. Particularly, we observe that in Figure~\ref{fig:423_comparison}, even though context-guided sequential editing fails to replace the trivet with a cocoa bean in the first edit, it can still correct its behavior in the subsequent edits and achieves a perfect edit after three steps. Overall, context-guided sequential editing effectively improves over single-turn editing and outperforms other sequential editing baselines.
    \item In %
    Figure~\ref{fig:2279_comparison}, we compare our methods against baselines on two dependent chains with 10 editing operations. %
    Context-guided sequential editing produces a high-quality result with only one incomplete edit (\ie, replaced the vase with a shoe but did not continue to update its material to fabric), whereas other proprietary baselines generate results with more incorrect edits.
    \item We provide additional qualitative results (Figures~\ref{fig:81_real_comparison}-\ref{fig:49_real_comparison}) on Complex-Edit~\cite{Yang2025ComplexEditCI} for sim-to-real generalization. In most of the real-world editing examples, we find that two-step context-guided sequential editing consistently demonstrates better instruction-following and identity preservation capabilities compared to other baselines. However, we also observe that sequential editing can sometimes fail to execute fine-grained edits (\eg,~\ref{fig:161_real_comparison}) or overly emphasize certain effects (\eg, Figures~\ref{fig:301_real_comparison},~\ref{fig:49_real_comparison}), resulting in compromised fidelity.
\end{itemize}

\begin{thebibliography}{44}
\providecommand{\natexlab}[1]{#1}
\providecommand{\url}[1]{\texttt{#1}}
\expandafter\ifx\csname urlstyle\endcsname\relax
  \providecommand{\doi}[1]{doi: #1}\else
  \providecommand{\doi}{doi: \begingroup \urlstyle{rm}\Url}\fi

\bibitem[Batifol et~al.(2025)Batifol, Blattmann, Boesel, Consul, Diagne, Dockhorn, English, English, Esser, Kulal, et~al.]{batifol2025flux}
Stephen Batifol, Andreas Blattmann, Frederic Boesel, Saksham Consul, Cyril Diagne, Tim Dockhorn, Jack English, Zion English, Patrick Esser, Sumith Kulal, et~al.
\newblock Flux. 1 kontext: Flow matching for in-context image generation and editing in latent space.
\newblock \emph{arXiv e-prints}, pages arXiv--2506, 2025.

\bibitem[Betker et~al.()Betker, Goh, Jing, TimBrooks, Wang, Li, LongOuyang, JuntangZhuang, JoyceLee, YufeiGuo, WesamManassra, PrafullaDhariwal, CaseyChu, YunxinJiao, and Ramesh]{BetkerImprovingIG}
James Betker, Gabriel Goh, Li Jing, † TimBrooks, Jianfeng Wang, Linjie Li, † LongOuyang, † JuntangZhuang, † JoyceLee, † YufeiGuo, † WesamManassra, † PrafullaDhariwal, † CaseyChu, † YunxinJiao, and Aditya Ramesh.
\newblock Improving image generation with better captions.

\bibitem[Brooks et~al.(2022)Brooks, Holynski, and Efros]{Brooks2022InstructPix2PixLT}
Tim Brooks, Aleksander Holynski, and Alexei~A. Efros.
\newblock Instructpix2pix: Learning to follow image editing instructions.
\newblock \emph{2023 IEEE/CVF Conference on Computer Vision and Pattern Recognition (CVPR)}, pages 18392--18402, 2022.

\bibitem[Cao et~al.(2023)Cao, Wang, Qi, Shan, Qie, and Zheng]{cao2023masactrl}
Mingdeng Cao, Xintao Wang, Zhongang Qi, Ying Shan, Xiaohu Qie, and Yinqiang Zheng.
\newblock Masactrl: Tuning-free mutual self-attention control for consistent image synthesis and editing.
\newblock In \emph{Proceedings of the IEEE/CVF international conference on computer vision}, pages 22560--22570, 2023.

\bibitem[Cao et~al.(2025)Cao, Zhang, Zheng, and Xia]{cao2025instruction}
Mingdeng Cao, Xuaner Zhang, Yinqiang Zheng, and Zhihao Xia.
\newblock Instruction-based image manipulation by watching how things move.
\newblock In \emph{Proceedings of the Computer Vision and Pattern Recognition Conference}, pages 2704--2713, 2025.

\bibitem[Chang et~al.(2025)Chang, Cao, Shi, Liu, Cai, Zhou, Huang, Wetzstein, Soleymani, and Wang]{Chang2025ByteMorphBI}
Di Chang, Mingdeng Cao, Yichun Shi, Bo Liu, Shengqu Cai, Shijie Zhou, Weilin Huang, Gordon Wetzstein, Mohammad Soleymani, and Peng Wang.
\newblock Bytemorph: Benchmarking instruction-guided image editing with non-rigid motions.
\newblock \emph{ArXiv}, abs/2506.03107, 2025.

\bibitem[Chen et~al.(2023)Chen, Huang, Liu, Shen, Zhao, and Zhao]{Chen2023AnyDoorZO}
Xi Chen, Lianghua Huang, Yu Liu, Yujun Shen, Deli Zhao, and Hengshuang Zhao.
\newblock Anydoor: Zero-shot object-level image customization.
\newblock \emph{2024 IEEE/CVF Conference on Computer Vision and Pattern Recognition (CVPR)}, pages 6593--6602, 2023.

\bibitem[Chen et~al.(2025)Chen, Zhang, Zhang, Zhou, Kim, Liu, Li, Zhang, Zhao, Wang, et~al.]{chen2025unireal}
Xi Chen, Zhifei Zhang, He Zhang, Yuqian Zhou, Soo~Ye Kim, Qing Liu, Yijun Li, Jianming Zhang, Nanxuan Zhao, Yilin Wang, et~al.
\newblock Unireal: Universal image generation and editing via learning real-world dynamics.
\newblock In \emph{Proceedings of the Computer Vision and Pattern Recognition Conference}, pages 12501--12511, 2025.

\bibitem[Couairon et~al.(2022)Couairon, Verbeek, Schwenk, and Cord]{couairon2022diffedit}
Guillaume Couairon, Jakob Verbeek, Holger Schwenk, and Matthieu Cord.
\newblock Diffedit: Diffusion-based semantic image editing with mask guidance.
\newblock \emph{arXiv preprint arXiv:2210.11427}, 2022.

\bibitem[Deng et~al.(2025)Deng, Zhu, Li, Gou, Li, Wang, Zhong, Yu, Nie, Song, et~al.]{deng2025emerging}
Chaorui Deng, Deyao Zhu, Kunchang Li, Chenhui Gou, Feng Li, Zeyu Wang, Shu Zhong, Weihao Yu, Xiaonan Nie, Ziang Song, et~al.
\newblock Emerging properties in unified multimodal pretraining.
\newblock \emph{arXiv preprint arXiv:2505.14683}, 2025.

\bibitem[Denninger et~al.(2023)Denninger, Winkelbauer, Sundermeyer, Boerdijk, Knauer, Strobl, Humt, and Triebel]{Denninger2023}
Maximilian Denninger, Dominik Winkelbauer, Martin Sundermeyer, Wout Boerdijk, Markus Knauer, Klaus~H. Strobl, Matthias Humt, and Rudolph Triebel.
\newblock Blenderproc2: A procedural pipeline for photorealistic rendering.
\newblock \emph{Journal of Open Source Software}, 8\penalty0 (82):\penalty0 4901, 2023.

\bibitem[Dong et~al.(2025)Dong, Chen, Lv, Yu, Zhang, Zhang, Zhu, Tian, Li, Moffatt, Christofferson, Fort, Pan, Yan, Wu, Ren, and Newcombe]{Dong_2025_CVPR}
Zhao Dong, Ka Chen, Zhaoyang Lv, Hong-Xing Yu, Yunzhi Zhang, Cheng Zhang, Yufeng Zhu, Stephen Tian, Zhengqin Li, Geordie Moffatt, Sean Christofferson, James Fort, Xiaqing Pan, Mingfei Yan, Jiajun Wu, Carl~Yuheng Ren, and Richard Newcombe.
\newblock Digital twin catalog: A large-scale photorealistic 3d object digital twin dataset.
\newblock In \emph{Proceedings of the IEEE/CVF Conference on Computer Vision and Pattern Recognition (CVPR)}, 2025.

\bibitem[Esser et~al.(2024)Esser, Kulal, Blattmann, Entezari, M\"{u}ller, Saini, Levi, Lorenz, Sauer, Boesel, Podell, Dockhorn, English, and Rombach]{pmlr-v235-esser24a}
Patrick Esser, Sumith Kulal, Andreas Blattmann, Rahim Entezari, Jonas M\"{u}ller, Harry Saini, Yam Levi, Dominik Lorenz, Axel Sauer, Frederic Boesel, Dustin Podell, Tim Dockhorn, Zion English, and Robin Rombach.
\newblock Scaling rectified flow transformers for high-resolution image synthesis.
\newblock In \emph{Proceedings of the 41st International Conference on Machine Learning}, pages 12606--12633. PMLR, 2024.

\bibitem[Fu et~al.(2023)Fu, Hu, Du, Wang, Yang, and Gan]{fu2023guiding}
Tsu-Jui Fu, Wenze Hu, Xianzhi Du, William~Yang Wang, Yinfei Yang, and Zhe Gan.
\newblock Guiding instruction-based image editing via multimodal large language models.
\newblock \emph{arXiv preprint arXiv:2309.17102}, 2023.

\bibitem[Haven()]{polyhaven}
Poly Haven.
\newblock \url{https://polyhaven.com/}.

\bibitem[Hertz et~al.(2022)Hertz, Mokady, Tenenbaum, Aberman, Pritch, and Cohen-Or]{hertz2022prompt}
Amir Hertz, Ron Mokady, Jay Tenenbaum, Kfir Aberman, Yael Pritch, and Daniel Cohen-Or.
\newblock Prompt-to-prompt image editing with cross attention control.
\newblock \emph{arXiv preprint arXiv:2208.01626}, 2022.

\bibitem[Huang et~al.(2024)Huang, Xie, Wang, Yuan, Cun, Ge, Zhou, Dong, Huang, Zhang, et~al.]{huang2024smartedit}
Yuzhou Huang, Liangbin Xie, Xintao Wang, Ziyang Yuan, Xiaodong Cun, Yixiao Ge, Jiantao Zhou, Chao Dong, Rui Huang, Ruimao Zhang, et~al.
\newblock Smartedit: Exploring complex instruction-based image editing with multimodal large language models.
\newblock In \emph{Proceedings of the IEEE/CVF Conference on Computer Vision and Pattern Recognition}, pages 8362--8371, 2024.

\bibitem[Kawar et~al.(2023)Kawar, Zada, Lang, Tov, Chang, Dekel, Mosseri, and Irani]{kawar2023imagic}
Bahjat Kawar, Shiran Zada, Oran Lang, Omer Tov, Huiwen Chang, Tali Dekel, Inbar Mosseri, and Michal Irani.
\newblock Imagic: Text-based real image editing with diffusion models.
\newblock In \emph{Proceedings of the IEEE/CVF conference on computer vision and pattern recognition}, pages 6007--6017, 2023.

\bibitem[Li et~al.(2025)Li, Wang, Yue, Cai, Liu, Fu, Guo, Zhu, Sharan, Jia, Neiswanger, Huang, Goldstein, and Goldblum]{Li2025ZebraCoTAD}
Ang Li, Charles~L. Wang, Kaiyu Yue, Zikui Cai, Ollie Liu, Deqing Fu, Peng Guo, Wang~Bill Zhu, Vatsal Sharan, Robin Jia, Willie Neiswanger, Furong Huang, Tom Goldstein, and Micah Goldblum.
\newblock Zebra-cot: A dataset for interleaved vision language reasoning.
\newblock \emph{ArXiv}, abs/2507.16746, 2025.

\bibitem[Meng et~al.(2021)Meng, Song, Song, Wu, Zhu, and Ermon]{meng2021sdedit}
Chenlin Meng, Yang Song, Jiaming Song, Jiajun Wu, Jun-Yan Zhu, and Stefano Ermon.
\newblock Sdedit: Image synthesis and editing with stochastic differential equations.
\newblock \emph{arXiv preprint arXiv:2108.01073}, 2021.

\bibitem[Nichol et~al.(2022)Nichol, Dhariwal, Ramesh, Shyam, Mishkin, Mcgrew, Sutskever, and Chen]{pmlr-v162-nichol22a}
Alexander~Quinn Nichol, Prafulla Dhariwal, Aditya Ramesh, Pranav Shyam, Pamela Mishkin, Bob Mcgrew, Ilya Sutskever, and Mark Chen.
\newblock {GLIDE}: Towards photorealistic image generation and editing with text-guided diffusion models.
\newblock In \emph{Proceedings of the 39th International Conference on Machine Learning}, pages 16784--16804. PMLR, 2022.

\bibitem[OpenAI(2025)]{openai2025addendum}
OpenAI.
\newblock Addendum to gpt-4o system card: Native image generation.
\newblock OpenAI, 2025.

\bibitem[Qian et~al.(2025)Qian, Bocek-Rivele, Song, Tong, Yang, Lu, Hu, and Gan]{qian2025picobanana400klargescaledatasettextguided}
Yusu Qian, Eli Bocek-Rivele, Liangchen Song, Jialing Tong, Yinfei Yang, Jiasen Lu, Wenze Hu, and Zhe Gan.
\newblock Pico-banana-400k: A large-scale dataset for text-guided image editing, 2025.

\bibitem[Qu et~al.(2025)Qu, Cheng, Yang, Zhao, Lin, Shi, Li, Wang, Chua, and Jiang]{Qu2025VINCIEUI}
Leigang Qu, Feng Cheng, Ziyan Yang, Qi Zhao, Shanchuan Lin, Yichun Shi, Yicong Li, Wenjie Wang, Tat-Seng Chua, and Lu Jiang.
\newblock Vincie: Unlocking in-context image editing from video.
\newblock \emph{ArXiv}, abs/2506.10941, 2025.

\bibitem[Ramesh et~al.(2021)Ramesh, Pavlov, Goh, Gray, Voss, Radford, Chen, and Sutskever]{pmlr-v139-ramesh21a}
Aditya Ramesh, Mikhail Pavlov, Gabriel Goh, Scott Gray, Chelsea Voss, Alec Radford, Mark Chen, and Ilya Sutskever.
\newblock Zero-shot text-to-image generation.
\newblock In \emph{Proceedings of the 38th International Conference on Machine Learning}, pages 8821--8831. PMLR, 2021.

\bibitem[Ramesh et~al.(2022)Ramesh, Dhariwal, Nichol, Chu, and Chen]{Ramesh2022HierarchicalTI}
Aditya Ramesh, Prafulla Dhariwal, Alex Nichol, Casey Chu, and Mark Chen.
\newblock Hierarchical text-conditional image generation with clip latents.
\newblock \emph{ArXiv}, abs/2204.06125, 2022.

\bibitem[Rombach et~al.(2021)Rombach, Blattmann, Lorenz, Esser, and Ommer]{Rombach2021HighResolutionIS}
Robin Rombach, A. Blattmann, Dominik Lorenz, Patrick Esser, and Bj{\"o}rn Ommer.
\newblock High-resolution image synthesis with latent diffusion models.
\newblock \emph{2022 IEEE/CVF Conference on Computer Vision and Pattern Recognition (CVPR)}, pages 10674--10685, 2021.

\bibitem[Saharia et~al.(2022)Saharia, Chan, Saxena, Li, Whang, Denton, Ghasemipour, Ayan, Mahdavi, Lopes, Salimans, Ho, Fleet, and Norouzi]{Saharia2022PhotorealisticTD}
Chitwan Saharia, William Chan, Saurabh Saxena, Lala Li, Jay Whang, Emily~L. Denton, Seyed Kamyar~Seyed Ghasemipour, Burcu~Karagol Ayan, Seyedeh~Sara Mahdavi, Raphael~Gontijo Lopes, Tim Salimans, Jonathan Ho, David~J. Fleet, and Mohammad Norouzi.
\newblock Photorealistic text-to-image diffusion models with deep language understanding.
\newblock \emph{ArXiv}, abs/2205.11487, 2022.

\bibitem[Sheynin et~al.(2023)Sheynin, Polyak, Singer, Kirstain, Zohar, Ashual, Parikh, and Taigman]{Sheynin2023EmuEP}
Shelly Sheynin, Adam Polyak, Uriel Singer, Yuval Kirstain, Amit Zohar, Oron Ashual, Devi Parikh, and Yaniv Taigman.
\newblock Emu edit: Precise image editing via recognition and generation tasks.
\newblock \emph{2024 IEEE/CVF Conference on Computer Vision and Pattern Recognition (CVPR)}, pages 8871--8879, 2023.

\bibitem[Sheynin et~al.(2024)Sheynin, Polyak, Singer, Kirstain, Zohar, Ashual, Parikh, and Taigman]{sheynin2024emu}
Shelly Sheynin, Adam Polyak, Uriel Singer, Yuval Kirstain, Amit Zohar, Oron Ashual, Devi Parikh, and Yaniv Taigman.
\newblock Emu edit: Precise image editing via recognition and generation tasks.
\newblock In \emph{Proceedings of the IEEE/CVF Conference on Computer Vision and Pattern Recognition}, pages 8871--8879, 2024.

\bibitem[Sim'eoni et~al.(2025)Sim'eoni, Vo, Seitzer, Baldassarre, Oquab, Jose, Khalidov, Szafraniec, Yi, Ramamonjisoa, Massa, Haziza, Wehrstedt, Wang, Darcet, Moutakanni, Sentana, Roberts, Vedaldi, Tolan, Brandt, Couprie, Mairal, J'egou, Labatut, and Bojanowski]{Simeoni2025DINOv3}
Oriane Sim'eoni, Huy~V. Vo, Maximilian Seitzer, Federico Baldassarre, Maxime Oquab, Cijo Jose, Vasil Khalidov, Marc Szafraniec, Seungeun Yi, Michael Ramamonjisoa, Francisco Massa, Daniel Haziza, Luca Wehrstedt, Jianyuan Wang, Timoth{\'e}e Darcet, Th{\'e}o Moutakanni, Leonel Sentana, Claire Roberts, Andrea Vedaldi, Jamie Tolan, John Brandt, Camille Couprie, Julien Mairal, Herv'e J'egou, Patrick Labatut, and Piotr Bojanowski.
\newblock Dinov3.
\newblock 2025.

\bibitem[Sun et~al.(2024)Sun, Cui, Zhang, Zhang, Yu, Wang, Rao, Liu, Huang, and Wang]{sun2024generative}
Quan Sun, Yufeng Cui, Xiaosong Zhang, Fan Zhang, Qiying Yu, Yueze Wang, Yongming Rao, Jingjing Liu, Tiejun Huang, and Xinlong Wang.
\newblock Generative multimodal models are in-context learners.
\newblock In \emph{Proceedings of the IEEE/CVF Conference on Computer Vision and Pattern Recognition}, pages 14398--14409, 2024.

\bibitem[Team(2025)]{gemini2025}
Google~Gemini Team.
\newblock Gemini 2.5 flash image (nano banana).
\newblock \url{https://gemini.google.com/}, 2025.

\bibitem[Textures()]{cctextures}
CC0 Textures.
\newblock \url{https://cc0-textures.com/}.

\bibitem[Tumanyan et~al.(2023)Tumanyan, Geyer, Bagon, and Dekel]{tumanyan2023plug}
Narek Tumanyan, Michal Geyer, Shai Bagon, and Tali Dekel.
\newblock Plug-and-play diffusion features for text-driven image-to-image translation.
\newblock In \emph{Proceedings of the IEEE/CVF conference on computer vision and pattern recognition}, pages 1921--1930, 2023.

\bibitem[Wei et~al.(2024)Wei, Xiong, Ren, Du, Zhang, and Chen]{Wei2024OmniEditBI}
Cong Wei, Zheyang Xiong, Weiming Ren, Xinrun Du, Ge Zhang, and Wenhu Chen.
\newblock Omniedit: Building image editing generalist models through specialist supervision.
\newblock \emph{ArXiv}, abs/2411.07199, 2024.

\bibitem[Wu et~al.(2025{\natexlab{a}})Wu, Li, Zhou, Lin, Gao, Yan, ming Yin, Bai, Xu, Chen, Chen, Tang, Zhang, Wang, Yang, Yu, Cheng, Liu, Li, Zhang, Meng, Wei, Ni, Chen, Cao, Peng, Qu, Wu, Wang, Yu, Wen, Feng, Xu, Wang, Zhang, Zhu, Wu, Cai, and Liu]{wu2025qwenimagetechnicalreport}
Chenfei Wu, Jiahao Li, Jingren Zhou, Junyang Lin, Kaiyuan Gao, Kun Yan, Sheng ming Yin, Shuai Bai, Xiao Xu, Yilei Chen, Yuxiang Chen, Zecheng Tang, Zekai Zhang, Zhengyi Wang, An Yang, Bowen Yu, Chen Cheng, Dayiheng Liu, Deqing Li, Hang Zhang, Hao Meng, Hu Wei, Jingyuan Ni, Kai Chen, Kuan Cao, Liang Peng, Lin Qu, Minggang Wu, Peng Wang, Shuting Yu, Tingkun Wen, Wensen Feng, Xiaoxiao Xu, Yi Wang, Yichang Zhang, Yongqiang Zhu, Yujia Wu, Yuxuan Cai, and Zenan Liu.
\newblock Qwen-image technical report, 2025{\natexlab{a}}.

\bibitem[Wu et~al.(2025{\natexlab{b}})Wu, Li, Zhou, Lin, Gao, Yan, ming Yin, Bai, Xu, Chen, Chen, Tang, Zhang, Wang, Yang, Yu, Cheng, Liu, mei Li, Zhang, Meng, Wei, Ni, Chen, Cao, Peng, Qu, Wu, Wang, Yu, Wen, Feng, Xu, Wang, Zhang, Zhu, Wu, Cai, and Liu]{Wu2025QwenImageTR}
Chenfei Wu, Jiahao Li, Jingren Zhou, Junyang Lin, Kaiyuan Gao, Kun Yan, Sheng ming Yin, Shuai Bai, Xiao Xu, Yilei Chen, Yuxiang Chen, Zecheng Tang, Zekai Zhang, Zhengyi Wang, An Yang, Bowen Yu, Chen Cheng, Da-Wei Liu, De mei Li, Hang Zhang, Hao Meng, Hu Wei, Ji-Li Ni, Kai Chen, Kuan Cao, Liang Peng, Lin Qu, Min Wu, Peng Wang, Shuting Yu, Tingkun Wen, Wensen Feng, Xiao-Xue Xu, Yi Wang, Yichang Zhang, Yong-An Zhu, Yujia Wu, Yu-Jiao Cai, and Ze-Yang Liu.
\newblock Qwen-image technical report.
\newblock \emph{ArXiv}, abs/2508.02324, 2025{\natexlab{b}}.

\bibitem[Xiao et~al.(2025)Xiao, Wang, Zhou, Yuan, Xing, Yan, Li, Wang, Huang, and Liu]{xiao2025omnigen}
Shitao Xiao, Yueze Wang, Junjie Zhou, Huaying Yuan, Xingrun Xing, Ruiran Yan, Chaofan Li, Shuting Wang, Tiejun Huang, and Zheng Liu.
\newblock Omnigen: Unified image generation.
\newblock In \emph{Proceedings of the Computer Vision and Pattern Recognition Conference}, pages 13294--13304, 2025.

\bibitem[Yang et~al.(2025)Yang, Hui, Zhao, Zhou, Ruiz, and Xie]{Yang2025ComplexEditCI}
Siwei Yang, Mude Hui, Bingchen Zhao, Yuyin Zhou, Nataniel Ruiz, and Cihang Xie.
\newblock Complex-edit: Cot-like instruction generation for complexity-controllable image editing benchmark.
\newblock \emph{ArXiv}, abs/2504.13143, 2025.

\bibitem[Yu et~al.(2024)Yu, Chow, Yue, Pan, Wu, Wan, Li, Tang, Zhang, and Zhuang]{yu2024anyedit}
Qifan Yu, Wei Chow, Zhongqi Yue, Kaihang Pan, Yang Wu, Xiaoyang Wan, Juncheng Li, Siliang Tang, Hanwang Zhang, and Yueting Zhuang.
\newblock Anyedit: Mastering unified high-quality image editing for any idea.
\newblock \emph{arXiv preprint arXiv:2411.15738}, 2024.

\bibitem[Zhang et~al.(2023)Zhang, Mo, Chen, Sun, and Su]{Zhang2023MagicBrushAM}
Kai Zhang, Lingbo Mo, Wenhu Chen, Huan Sun, and Yu Su.
\newblock Magicbrush: A manually annotated dataset for instruction-guided image editing.
\newblock \emph{ArXiv}, abs/2306.10012, 2023.

\bibitem[Zhao et~al.(2024)Zhao, Ma, Chen, Si, Wu, An, Yu, Zhang, Li, and Chang]{Zhao2024UltraEditIF}
Haozhe Zhao, Xiaojian Ma, Liang Chen, Shuzheng Si, Rujie Wu, Kaikai An, Peiyu Yu, Minjia Zhang, Qing Li, and Baobao Chang.
\newblock Ultraedit: Instruction-based fine-grained image editing at scale.
\newblock \emph{ArXiv}, abs/2407.05282, 2024.

\bibitem[Zhou et~al.(2025)Zhou, Deng, He, Dong, and Tang]{Zhou2025MultiturnCI}
Zijun Zhou, Yingying Deng, Xiangyu He, Weiming Dong, and Fan Tang.
\newblock Multi-turn consistent image editing.
\newblock \emph{ArXiv}, abs/2505.04320, 2025.

\end{thebibliography}
\end{document}